\documentclass[10pt,twocolumn,letterpaper]{article}
\usepackage[pagenumbers]{cvpr}
\usepackage[T1]{fontenc}
\usepackage{graphicx}
\usepackage{amsmath,amssymb}
\usepackage{booktabs}
\usepackage{xspace}
\usepackage{tikz}
\usetikzlibrary{arrows.meta,positioning,fit,backgrounds}
\graphicspath{{figures/}}
\newcommand{\method}{BodyWeave\xspace}
\newcommand{\dataset}{GarmentWeave3D\xspace}
\usepackage{longtable}
\usepackage{array}
\usepackage{colortbl}
\definecolor{tablegood}{RGB}{204,255,204}
\definecolor{tablemid}{RGB}{255,255,204}
\definecolor{tablepoor}{RGB}{255,204,204}
\newcommand{\metriccell}[2]{\cellcolor{#1}#2}
\newcommand{\tableformat}{\renewcommand{\arraystretch}{1.12}\setlength{\aboverulesep}{1.8pt}\setlength{\belowrulesep}{1.8pt}\setlength{\heavyrulewidth}{0.8pt}\setlength{\lightrulewidth}{0.4pt}}

\definecolor{cvprblue}{rgb}{0.21,0.49,0.74}
\usepackage[breaklinks,colorlinks,allcolors=cvprblue]{hyperref}
\def\confName{CVPR}
\def\confYear{2027}
\title{BodyWeave: Native Body-Conditioned 3D Garment Generation}
\author{Junyi Zhang\textsuperscript{1}\quad Yuanchuan Wang\textsuperscript{1}\quad Qichao Wang\textsuperscript{1}\\
Wenzhe Qian\textsuperscript{1}\quad Shizhuo Cheng\textsuperscript{1}\quad Min Zhang\textsuperscript{1}\\
\textsuperscript{1}Zhejiang University\\
\href{mailto:gavin.jy@zju.edu.cn}{\texttt{gavin.jy@zju.edu.cn}}}
\makeatletter
\patchcmd{\@maketitle}{\vskip .375in}{\vskip 20pt}{}{}
\patchcmd{\@maketitle}{\vspace*{24pt}}{\vspace*{10pt}}{}{}
\apptocmd{\@maketitle}{\vspace{-24pt}\begin{minipage}{\textwidth}
\centering
\setlength{\abovecaptionskip}{0.5em}
\setlength{\belowcaptionskip}{1em}
\includegraphics[width=\linewidth]{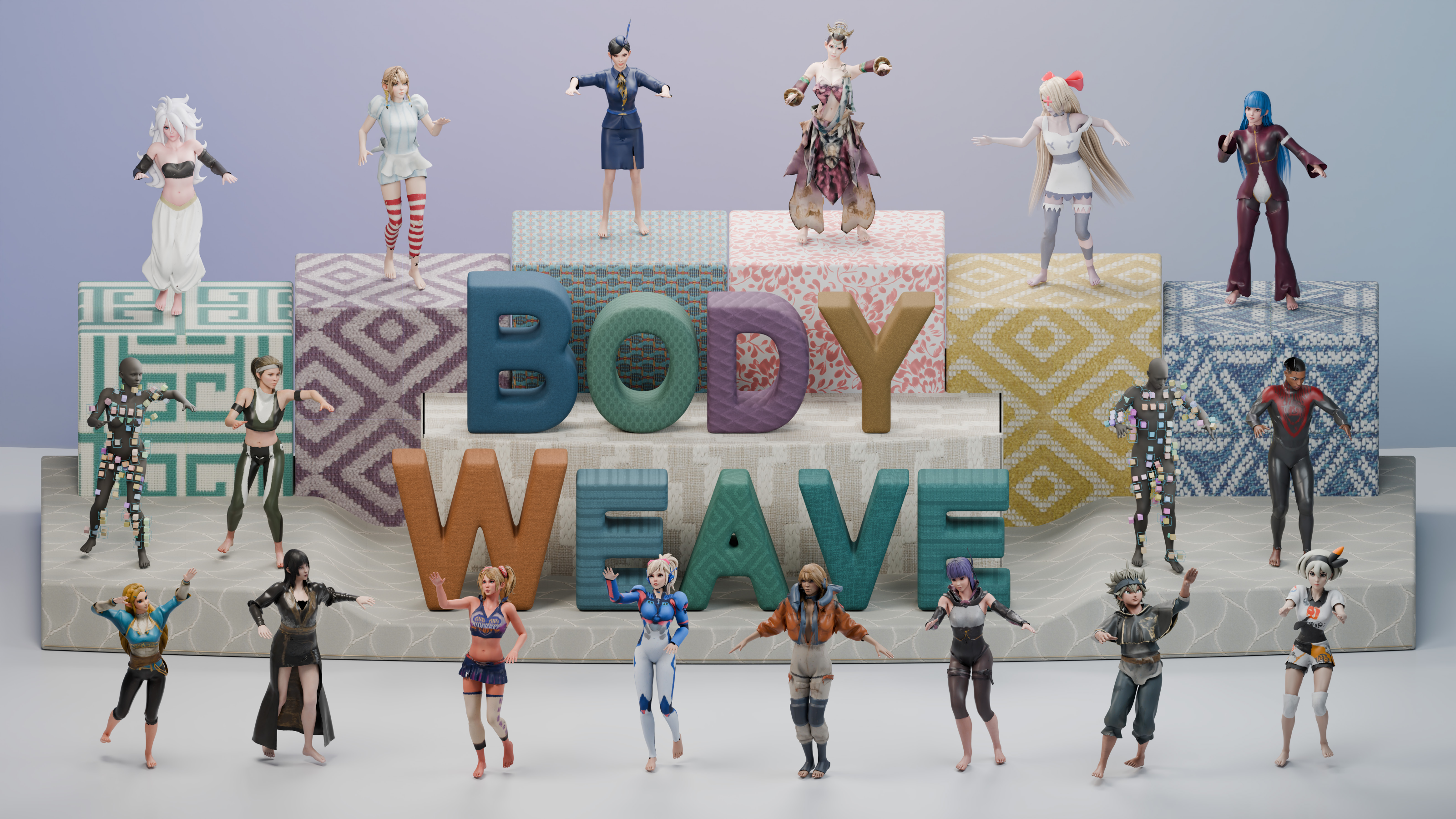}
\captionof{figure}{\textbf{\method generates garments with native 3D body conditioning.} Given reference images and a target body, our method generates garment geometry directly in the body's coordinate frame together with PBR appearance, as illustrated by results on \textbf{Artist-Created Garments}. The dark gray bodies on either side depict the input body conditions. The pose-aligned colored cells visualize first-stage garment support, marking occupied locations for subsequent shape denoising, with colors encoding latent features. The lettering and display pillars are textured with fabric materials from \textbf{\dataset}.}
\label{fig:teaser}
\end{minipage}
\par
}{}{}
\makeatother
\begin{document}
\maketitle
\begin{abstract}
Generating 3D garments for digital characters requires preserving a reference design while adapting the garment to the target body's shape and pose.
Sewing-pattern methods can struggle to reproduce unfamiliar designs and fine details, while image-to-3D generators offer rich shape priors but typically lack explicit 3D body context.
We present \textbf{BodyWeave}, a framework that unifies garment generation and fitting through native 3D body conditioning.
Building on 3D generative priors, BodyWeave uses the body as a spatial reference throughout geometry generation: a \textbf{Global Body Adapter} guides garment placement and extent in the shared coordinate frame, while \textbf{Local Body Attention} guides surface refinement using nearby body geometry.
A \textbf{Surface Penetration Loss} supervises decoded surfaces to reduce body intersections, and the resulting geometry conditions material synthesis to produce garments with physically based rendering attributes.
To support learning across designs and wearers, we construct \textbf{GarmentWeave3D} with 293K aligned garment--body observations, including 114K same-design cross-body pairs.
Evaluations on out-of-domain \textbf{Artist-Created Garments} demonstrate high geometric fidelity and low body penetration.
Cross-body evaluations demonstrate accurate adaptation to new wearers with consistent garment--body spacing, while appearance comparisons show faithful fabric reproduction.
Additional results and resources are available at \url{https://gavin-jy.github.io/bodyweave-page/}.
\par\end{abstract}

\section{Introduction}
\label{sec:intro}
High-quality 3D garments are fundamental to digital characters in entertainment, virtual reality, and fashion.
Given reference images and a target body, garment generation must preserve the intended design while accommodating the wearer's shape and pose.
The images specify garment structure and appearance, while the body provides the geometric context in which the garment must fit.

Existing approaches offer complementary strengths in design reconstruction and body fitting.
Sewing-pattern methods recover panels and stitching relationships from visual or textual inputs, typically using simulation to dress the target body~\cite{liu2023towards,he2024dresscode,nakayama2025aipparel,bian2025chatgarment,liu2025multimodal}.
Their construction structure supports fitting and editing, while pattern representations and training coverage can constrain the recovery of unfamiliar designs and fine details.
Large-scale 3D generative priors provide flexible surface representations for diverse garment geometry~\cite{xiang2025structured,xiang2026native}.
Garment3DGen uses generated shape guidance to deform a garment template~\cite{sarafianos2025garment3dgen}.
To achieve body compatibility, BAG guides multi-view image generation with body projections and subsequently corrects the generated geometry~\cite{luo2026bag}, while Tailor constrains text-guided template deformation using body geometry~\cite{sun2026text}.
However, their generative priors lack direct 3D body conditioning, leaving placement and local body compatibility to image-space guidance or geometric optimization.

\textit{Our insight is that fitting can be learned within geometry generation using the body as a spatial reference at two scales.}
In a shared 3D coordinate frame, the body's overall configuration guides garment placement and extent, while nearby surface geometry guides local shape.
We propose \textbf{\method}, a framework that integrates image guidance and spatial body conditioning to preserve garment design while adapting geometry to the target wearer.

Building on pretrained 3D generative priors~\cite{xiang2026native}, \method incorporates hierarchical body conditioning into spatial structure generation and surface refinement.
A \textbf{Global Body Adapter} (GBA) injects aligned body features into the sparse-structure transformer to predict garment placement and extent.
On this predicted support, \textbf{Local Body Attention} (LBA) connects garment features to nearby body geometry during surface refinement.
A \textbf{Surface Penetration Loss} supervises decoded garment surfaces during training, complementing latent-space generation objectives with explicit constraints on body intersections.
The refined geometry and reference images then condition physically based rendering (PBR) material synthesis, yielding a body-aligned garment mesh with appearance, without post-generation alignment (\cref{fig:teaser}).

To support learning both detailed garment geometry and design-preserving body adaptation, we construct \mbox{\textbf{\dataset}} by combining public synthetic garment data~\cite{korosteleva2021generating,korosteleva2024garmentcodedata} with our curated high-detail garment assets.
The dataset contains \textit{293K} aligned garment--body observations, including \textit{114K} same-design cross-body pairs.
These pairs hold the source garment images fixed while changing the target body and corresponding garment geometry, supervising adaptation that preserves design identity.
We enrich the observations with approximately \textit{2.5K} curated fabric materials, using shared material assignments across paired bodies to preserve appearance identity.

Experiments demonstrate \method's advantages in geometric fidelity, body adaptation, and fabric appearance.
On \textbf{GarmentWeave3D} and the out-of-domain \textbf{Artist-Created Garments} collection, \method achieves the lowest mean Chamfer distance among the compared garment-generation systems.
On the latter, which contains complex silhouettes and fine geometric details, it achieves a mean Chamfer distance of \textit{16.14\,mm} with \textit{0.86\%} body penetration.
Cross-body evaluations show accurate adaptation to new wearers with consistent garment--body spacing, while comparisons on shared geometry demonstrate faithful fabric appearance in both input and held-out views.

Our main contributions are summarized as follows:
\begin{itemize}
  \item We propose \textbf{\method}, a native 3D body-conditioned framework that unifies image-guided garment generation and target-body fitting.
  \item We introduce \textbf{GBA} and \textbf{LBA} for hierarchical body conditioning, together with decoded-surface penetration supervision for body--garment compatibility.
  \item We construct \textbf{\dataset} with aligned observations and same-design cross-body pairs, and demonstrate accurate reconstruction, body adaptation, and appearance synthesis on in-domain and artist-created garments.
\end{itemize}

\begin{figure*}[!t]
\centering
\includegraphics[width=\textwidth]{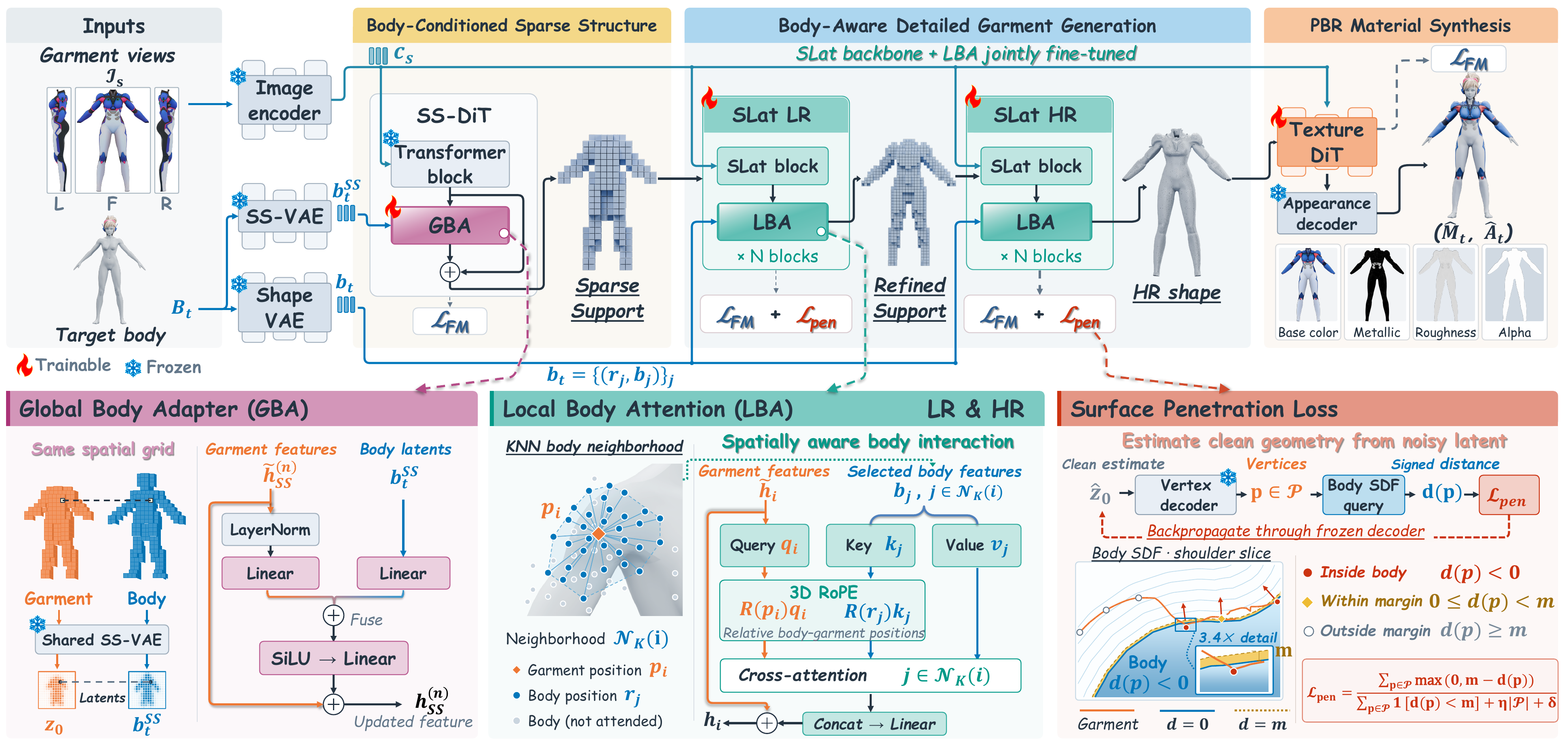}
\caption{\textbf{BodyWeave pipeline.} Given three garment views and a target body, BodyWeave generates garment geometry in the body's coordinate frame. GBA predicts body-aligned sparse support to establish global placement. LBA then injects local body geometry into low- and high-resolution shape generation, with surface penetration supervision regularizing both stages.}
\label{fig:pipeline}
\end{figure*}

\section{Related Work}
\label{sec:related}

\paragraph{Sewing-pattern generation.}
SewFormer, DressCode, AIpparel, ChatGarment, and SewingLDM recover or generate sewing patterns from images, text, and multimodal inputs~\cite{liu2023towards,he2024dresscode,nakayama2025aipparel,bian2025chatgarment,liu2025multimodal}.
Dress-1-to-3 refines patterns through diffusion-guided differentiable simulation~\cite{li2025dress}, while Garment Particles couples 2D patterns and 3D garments in a unified representation~\cite{nakayama2026garment}.
Despite these advances, supported pattern representations and training coverage can limit geometric diversity and fidelity to reference images, particularly for unfamiliar silhouettes and fine structural details.
\method uses pretrained 3D surface priors and aligned garment--body supervision to recover detailed geometry while adapting it to the wearer.
Direct surface generation avoids requiring image details to be recovered as panels and seams, supporting the geometric fidelity observed in our reconstruction evaluations.
\vspace{-1em}
\paragraph{Image-guided 3D garment generation.}
Garment3DGen and GarmentDreamer use generated 3D guidance to deform garment meshes~\cite{sarafianos2025garment3dgen,li2025garmentdreamer}, while GarmentCrafter improves garment reconstruction through progressive novel-view synthesis~\cite{wang2026garmentcrafter}.
Large-scale models such as CLAY and TRELLIS provide transferable surface priors~\cite{zhang2024clay,xiang2025structured,xiang2026native}.
However, image-conditioned shape generation alone does not specify how the garment should fit a separately supplied target body.
Visual reconstruction therefore does not ensure the correct body-relative placement or local surface compatibility.
\method supplies body features during both sparse-support prediction and surface refinement, allowing the same image-defined design to generate different geometry for different wearers in their coordinate frames.
\vspace{-1em}
\paragraph{Body alignment in generative pipelines.}
SMPLicit and Design2Cloth establish body-conditioned clothing representations~\cite{corona2021smplicit,zheng2024design2cloth}.
Integrating this body context into pretrained generative priors requires control over both garment placement and local fit.
BAG supplies body projections to multi-view image generation, followed by 3D generation, alignment, and penetration handling~\cite{luo2026bag}.
Tailor uses body-aware geometric constraints during text-guided template deformation~\cite{sun2026text}.
\method instead conditions 3D garment generation directly on the target body, learning garment placement, local fit, and shape together.
This makes body geometry available throughout generation to guide surface placement and constrain intersections, supporting more accurate fitting and better penetration control.
Supplementary Section~\ref{sec:supp-background} provides further background on garment representations and body-aware generation.

\section{Method}
\label{sec:method}
\begin{figure*}[!t]
\centering
\includegraphics[width=0.9\textwidth,pagebox=cropbox]{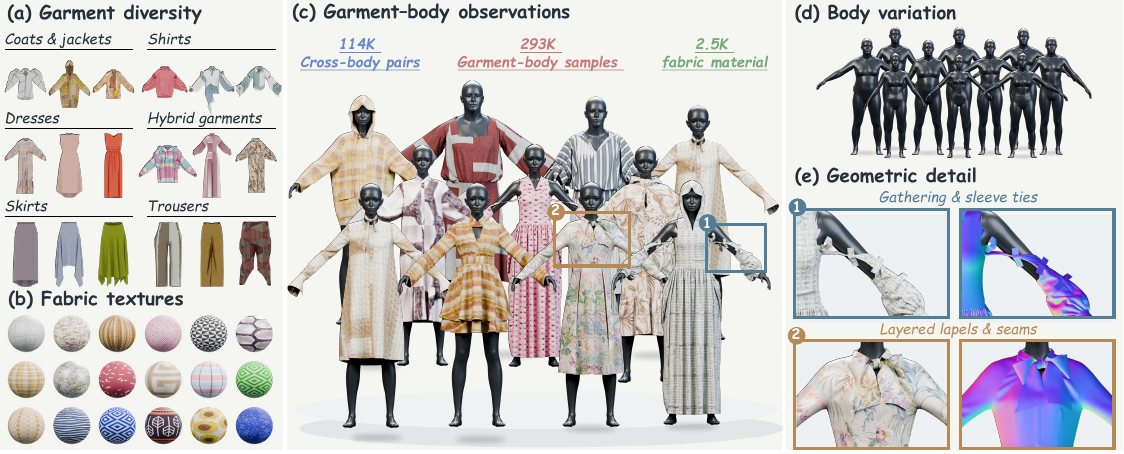}
\caption{\textbf{GarmentWeave3D.} (a) Garment category diversity. (b) Fabric materials visualized on spheres. (c) Simulated garments from GarmentWeave3D-Detail with assigned PBR materials, alongside dataset-scale statistics. (d) Body shape variation. (e) Gathering and sleeve ties, and layered lapels and seams from the numbered regions in (c), shown with paired textured and surface-normal renderings.}
\label{fig:dataset}
\end{figure*}
Given orthographic front, left, and right views $\mathcal I_s=\{I_s^F,I_s^L,I_s^R\}$ and a target 3D body $B_t$, \method generates a garment mesh and its PBR appearance:
\vspace{-0.5em}
\begin{equation}
  (\hat M_t,\hat A_t)=F(\mathcal I_s,B_t),
  \label{eq:task}
  \vspace{-0.5em}
\end{equation}
where $F$ denotes the full generation pipeline, $\hat M_t$ is a garment mesh in the body's coordinate frame, and $\hat A_t$ denotes its PBR attributes.
As illustrated in \cref{fig:pipeline}, we design a coarse-to-fine framework that conditions garment geometry on the target body and synthesizes materials on the generated surface.
We construct \textbf{\dataset} with aligned garment--body pairs, fabric materials, and multiview renders to supervise geometry, body adaptation, and appearance (\cref{sec:data}).
We use GBA to predict garment placement from image features $c_s$ extracted from $\mathcal I_s$ and target-body geometry (\cref{sec:ss}).
We refine the surface through LBA and penetration supervision, and generate PBR attributes conditioned on geometry and reference images (\cref{sec:geometry}).

\subsection{Simulation-Grounded Garment--Body Pairing}
\label{sec:data}
We construct \textbf{\dataset} (\cref{fig:dataset}) to support detailed design reconstruction, body adaptation, and fabric appearance within a shared training collection.
The Sewing-Pattern Dataset~\cite{korosteleva2021generating} provides simulated garments covering varied designs.
GarmentCodeData (v2)~\cite{korosteleva2024garmentcodedata} contributes 114.7K same-design cross-body pairs to supervise body adaptation.
We hold source garment images fixed while changing the target body and corresponding garment geometry, teaching body-dependent shape changes that preserve the reference design.
To extend geometric detail beyond these sources, we curate high-detail assets with dense gathers and layered structures.
The combined collection contains 293.8K observations, preserving simulated spatial relationships in a shared coordinate frame and physical scale.

Geometry alone does not provide the fabric appearance needed for image-conditioned generation and PBR synthesis.
We assign curated fabric materials and render garment-only orthographic views to supply both visual design references and appearance supervision.
Shared material assignments across paired bodies keep appearance consistent as geometry changes, allowing the body condition to specify the wearer while the images specify the garment.
Data-source statistics, material selection, and rendering details are provided in Supplementary Section~\ref{sec:dataset-construction}.

\subsection{Body-Conditioned Sparse Structure}
\label{sec:ss}
We adopt the sparse structure and structured latent representations of TRELLIS.2~\cite{xiang2026native} to generate garments in a coarse-to-fine manner.
The sparse structure specifies occupied spatial locations, while structured latents encode local surface geometry at those locations.
We first predict garment support from the reference images and target body, establishing where subsequent shape denoising takes place.
Conditioning this support on the body allows garment placement and extent to be determined before detailed surface generation.

To learn body-dependent support while retaining the pretrained shape prior, we freeze the sparse-structure diffusion transformer (SS-DiT) and introduce GBA after each transformer block.
This concentrates learning in body-conditioned residuals, allowing the model to acquire garment--body relationships from paired data while retaining the pretrained backbone parameters.

\vspace{-1em}
\paragraph{Global Body Adapter.}
The same frozen sparse-structure VAE encodes garment and body occupancies into spatially aligned latent grids of resolution $N_{\mathrm{SS}}^3$, where $N_{\mathrm{SS}}$ denotes the resolution along each axis.
The body tokens $b_t^{\mathrm{SS}}$ remain unnoised and provide a fixed spatial reference throughout garment denoising.
At flow time $\tau$, the $n$-th pretrained transformer block $\mathcal T_n$ maps its input features $h^{(n-1)}$ to $\widetilde h^{(n)}=\mathcal T_n(h^{(n-1)},\tau,c_s)$.
GBA then injects the aligned body features through a residual update
\begin{align}
  a^{(n)} &=
  P_n^h\!\left(\operatorname{LN}(\widetilde h^{(n)})\right)
  +P_n^b(b_t^{\mathrm{SS}}), \nonumber\\
  h^{(n)} &= \widetilde h^{(n)}+
  P_n^o\!\left(\operatorname{SiLU}(a^{(n)})\right),
  \label{eq:ss}
\end{align}
where $a^{(n)}$ is the fused adapter feature and $h^{(n)}$ is the output after body injection.
The learned affine projections $P_n^h$ and $P_n^b$ map garment and body features into the adapter space, while $P_n^o$ projects the fused features back to the transformer feature space.
Here $\operatorname{LN}$ denotes layer normalization.
Spatial alignment allows each garment token to incorporate body information at the corresponding location, guiding garment placement and extent.
Zero-initializing $P_n^o$ preserves the pretrained mapping at initialization.
\vspace{-1em}
\paragraph{Training objective.}
We train GBA with conditional flow matching.
Given a clean SS latent $z_0$, we construct $z_\tau=(1-\tau)z_0+\beta_\tau\epsilon$, where $\epsilon\sim\mathcal N(0,I)$, $\tau\in[0,1]$, and $\beta_\tau=\sigma+(1-\sigma)\tau$.
The velocity prediction $\hat v=v_\theta(z_\tau,\tau,c_s,b_t^{\mathrm{SS}})$ is supervised by
\begin{equation}
  \mathcal L_{\mathrm{FM}}
  =\mathbb E
  \!\left[\frac{1}{D}\left\|\hat v-\bigl((1-\sigma)\epsilon-z_0\bigr)\right\|_2^2\right],
  \label{eq:fm}
\end{equation}
where $D$ counts latent scalar entries in the batch and $\sigma$ is the residual noise level.
Sampling settings are provided in Supplementary Section~\ref{sec:method-details}.

\subsection{Body-Aware Detailed Garment Generation}
\label{sec:geometry}
\label{sec:appearance}
With garment placement established by the SS support, we recover detailed surface geometry through low-resolution (LR) and high-resolution (HR) shape generation.
Both stages generate shape latents on progressively refined sparse supports, using local body conditioning and decoded-surface supervision to guide geometric adaptation.
The resulting geometry then conditions PBR material synthesis to complete the garment asset.
\vspace{-1em}
\paragraph{Local Body Attention.}
Local garment shape depends on nearby body geometry, motivating spatially aware interactions between their distinct sparse supports.
The same frozen shape VAE encodes garment and body geometry in a shared coordinate frame at the same grid resolution.
A learned linear layer projects body features to $b_j$ at positions $r_j$, forming the body condition $b_t=\{(r_j,b_j)\}_j$.
For each garment position $p_i$, LBA selects the $K$ nearest body tokens by Euclidean distance, forming $\mathcal N_K(i)$.

We insert LBA after every structured-latent (SLat) transformer block in both shape stages.
Given the garment feature $\widetilde h_i$ before body injection, the residual update is
\begin{equation}
  h_i=\widetilde h_i+P_o\!\left(
  \operatorname{Attn}_{j\in\mathcal N_K(i)}
  \!\bigl(R(p_i)q_i,R(r_j)k_j,v_j\bigr)\right).
  \label{eq:local}
\end{equation}
Here $q_i$ is a learned projection of the layer-normalized garment feature, $k_j,v_j$ are learned projections of $b_j$, and $\operatorname{Attn}$ denotes standard multi-head scaled dot-product attention with output projection $P_o$.
The 3D rotary position encoding $R$ captures relative body--garment positions, while the KNN restriction focuses attention on nearby body geometry and reduces body-attention interactions.
Garment self-attention propagates this local context across the garment.
We jointly fine-tune the SLat backbone and body modules, with $P_o$ initialized to zero.
\vspace{-1em}
\paragraph{Geometry supervision.}
The LR and HR shape generators are trained separately using the flow-matching objective in \cref{eq:fm}, each with its own velocity network and shape latents.
Here, $z_0$ denotes the garment shape latent and $b_t$ the body position--feature pairs, with $\hat v=v_\theta(z_\tau,\tau,c_s,b_t)$; both the SLat backbone and LBA are optimized.

Our \textbf{Surface Penetration Loss} directly constrains body intersections in decoded geometry, which latent flow matching does not explicitly measure.
Using the body's signed distance field (SDF), it penalizes penetrating vertices and those within a small safety margin $m>0$ outside the body.
The penalty becomes inactive once the margin is reached, reducing intersections without exerting persistent outward pressure on the garment.
The predicted velocity yields the clean-latent estimate $\hat z_0=(1-\sigma)z_\tau-\beta_\tau\hat v$, which the frozen, differentiable shape decoder maps to garment vertex positions.
Let $\mathcal P$ collect the sampled vertices used for surface supervision across the current batch.
For each $p\in\mathcal P$, let $d(p)$ be the signed distance to its corresponding target body, positive outside.
The loss measures violations as
\begin{align}
  \rho(p)&=\max\!\left(0,m-d(p)\right), \nonumber\\
  \mathcal L_{\mathrm{pen}}
  &=
  \frac{\sum_{p\in\mathcal P}\rho(p)}
  {\sum_{p\in\mathcal P}\mathbf 1[\rho(p)>0]
   +\eta|\mathcal P|+\delta},
  \label{eq:surface}
\end{align}
Here $\rho(p)$ denotes the margin violation at point $p$.
Normalization by the violating-point count keeps sparse violations from being diluted by nonviolating vertices.
The positive constants $\eta$ and $\delta$ stabilize the denominator when few or no vertices violate the margin.

The total objective for each shape stage combines the two terms as
\begin{equation}
  \mathcal L_{\mathrm{shape}}
  =\mathcal L_{\mathrm{FM}}+\mathcal L_{\mathrm{pen}},
  \label{eq:shape-objective}
\end{equation}
The penetration term backpropagates through the frozen decoder to the shape generator, complementing latent flow matching with explicit surface supervision.
Representation and training settings are detailed in Supplementary~\ref{sec:method-details}.
\vspace{-1em}
\paragraph{PBR material synthesis.}
To equip the body-adapted garment with appearance consistent with the reference images, we generate PBR attributes conditioned on the images and refined geometry.
Following the geometry-conditioned appearance architecture of TRELLIS.2~\cite{xiang2026native}, the generator concatenates noisy appearance features with co-located shape features and is trained with conditional flow matching.
Training uses ground-truth geometry, while inference uses the generated HR shape.
An appearance decoder maps the generated latents to base color, metallic, roughness, and alpha attributes on the garment surface.

\section{Experiments}
\tableformat
\label{sec:experiments}

\begin{figure*}[!tbp]
\centering
\includegraphics[width=0.95\textwidth]{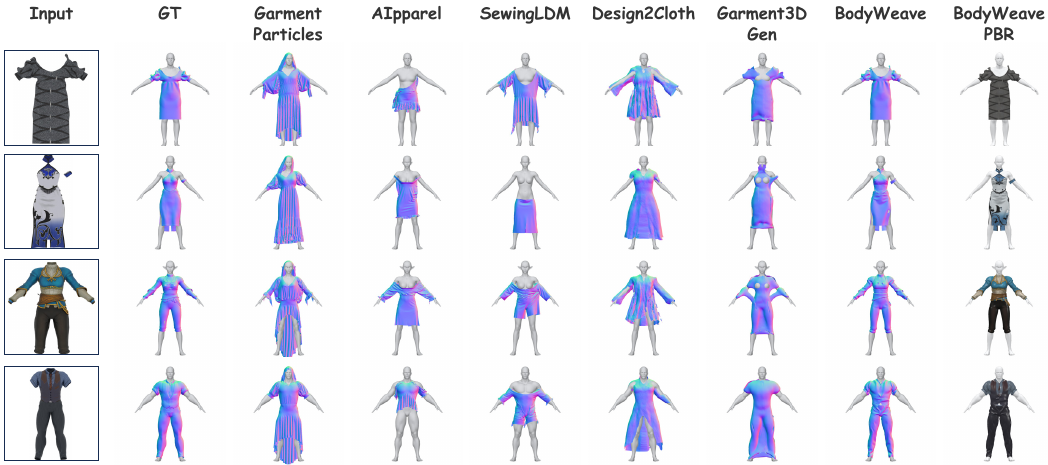}
\caption{\textbf{Geometric reconstruction comparisons on GarmentWeave3D and Artist-Created Garments}. Each row shows the input, ground truth, and reconstruction results. The final column presents our PBR appearance. Geometry is visualized with normal rendering.}
\label{fig:reconstruction}
\end{figure*}

\subsection{Experimental Setup}
\paragraph{Datasets.}
We train on \textbf{\dataset} (\cref{sec:data}) and evaluate reconstruction on 64 garment--body observations sampled from the test splits of its three constituent datasets.
We also evaluate on \textbf{Artist-Created Garments}, an out-of-domain collection of 49 assets with complex silhouettes and fine detail.
Cross-body evaluation uses eight same-design, same-material pairs from GarmentCodeData (v2)~\cite{korosteleva2024garmentcodedata} in both directions, yielding 16 tasks that retain source images while changing the target body.
\vspace{-1em}
\paragraph{Implementation details.}
\label{sec:implementation}
We initialize the generation stages from TRELLIS.2~\cite{xiang2026native}, training GBA with the SS backbone frozen and jointly fine-tuning the SLat backbone and body modules in LR/HR.
HR is initialized from the trained LR model.
The main SS, LR, HR, and PBR checkpoints use 20K, 10K, 65K, and 58K optimization steps, respectively.
LR/HR training combines flow matching and Surface Penetration Loss with a safety margin of $2.32\,\mathrm{mm}$.
More details are provided in Supplementary Section~\ref{sec:method-details}.
\vspace{-1em}
\paragraph{Baselines.}
We compare BodyWeave with state-of-the-art (SOTA) garment generation methods spanning simulation-based and direct 3D approaches: Garment Particles~\cite{nakayama2026garment}, AIpparel~\cite{nakayama2025aipparel}, SewingLDM~\cite{liu2025multimodal}, Design2Cloth~\cite{zheng2024design2cloth}, and Garment3DGen~\cite{sarafianos2025garment3dgen}.
We retain each method's native input interface and evaluate its complete generation and body adaptation pipeline.
Garment Particles, AIpparel, and SewingLDM generate garments followed by simulation on the target body.
Design2Cloth uses target-body pose transfer.
Garment3DGen uses ground-truth-assisted placement for reconstruction and fixed body-relative placement for cross-body evaluation.
For appearance synthesis, we compare with FabricDiffusion~\cite{zhang2024fabricdiffusion} and TRELLIS.2~\cite{xiang2026native} on shared geometry.
Body simulation and baseline protocols are detailed in Supplementary Section~\ref{sec:supp-protocol}.

\begin{figure*}[!tbp]
\centering
\includegraphics[width=0.95\textwidth]{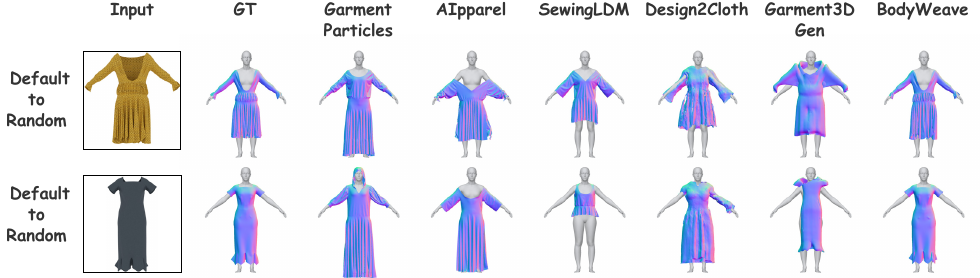}
\caption{\textbf{Cross-body garment adaptation} from the default body to a randomly sampled target body. All methods retain the source garment images while adapting the garment to the new body. Ground truth is obtained by simulating the same garment on the target body.}
\label{fig:adaptation}
\end{figure*}

\begin{table*}[!tbp]
\caption{Geometric reconstruction fidelity and body compatibility on GarmentWeave3D and Artist-Created Garments. \textbf{Bold} and \underline{underlined} values mark the best and second-best results. Colors rank each column from {\setlength{\fboxsep}{1pt}\colorbox{tablegood}{green}} (best) through {\setlength{\fboxsep}{1pt}\colorbox{tablemid}{yellow}} to {\setlength{\fboxsep}{1pt}\colorbox{tablepoor}{red}} (worst).}
\label{tab:reconstruction}
\centering\footnotesize\setlength{\tabcolsep}{1.3pt}\renewcommand{\arraystretch}{1.15}
\begin{tabular*}{\textwidth}{@{}>{\raggedright\arraybackslash}p{\dimexpr0.15\textwidth-\tabcolsep\relax}*{12}{>{\raggedleft\arraybackslash}p{\dimexpr0.0708333\textwidth-2\tabcolsep\relax}}@{\hspace{\tabcolsep}}}
\toprule
 & \multicolumn{6}{c}{\textbf{GarmentWeave3D}} & \multicolumn{6}{c}{\textbf{Artist-Created}} \\
\cmidrule(lr){2-7}\cmidrule(lr){8-13}
\textbf{Method} & \textbf{CD$\downarrow$} & \textbf{F@10$\uparrow$} & \textbf{Normal$\uparrow$} & \textbf{Pen.$\downarrow$} & \textbf{P95$\downarrow$} & {\scriptsize\textbf{Succ.\%$\uparrow$}} & \textbf{CD$\downarrow$} & \textbf{F@10$\uparrow$} & \textbf{Normal$\uparrow$} & \textbf{Pen.$\downarrow$} & \textbf{P95$\downarrow$} & {\scriptsize\textbf{Succ.\%$\uparrow$}} \\
\midrule
Garment Particles~\cite{nakayama2026garment} & \metriccell{tablegood!20!tablemid}{97.39} & \metriccell{tablepoor!60!tablemid}{26.81} & \metriccell{tablepoor!100!tablemid}{0.687} & \metriccell{tablepoor!20!tablemid}{0.760} & \metriccell{tablepoor!20!tablemid}{14.64} & \metriccell{tablepoor!100!tablemid}{71.88\%} & \metriccell{tablegood!20!tablemid}{61.84} & \metriccell{tablepoor!20!tablemid}{26.64} & \metriccell{tablepoor!100!tablemid}{0.666} & \metriccell{tablegood!60!tablemid}{\underline{0.662}} & \metriccell{tablepoor!20!tablemid}{25.07} & \metriccell{tablegood!33!tablemid}{75.51\%} \\
AIpparel~\cite{nakayama2025aipparel} & \metriccell{tablepoor!60!tablemid}{120.06} & \metriccell{tablegood!20!tablemid}{32.63} & \metriccell{tablepoor!20!tablemid}{0.715} & \metriccell{tablegood!20!tablemid}{0.465} & \metriccell{tablegood!20!tablemid}{8.86} & \metriccell{tablepoor!33!tablemid}{85.94\%} & \metriccell{tablepoor!60!tablemid}{117.23} & \metriccell{tablegood!20!tablemid}{30.40} & \metriccell{tablepoor!60!tablemid}{0.689} & \metriccell{tablepoor!20!tablemid}{2.259} & \metriccell{tablepoor!60!tablemid}{30.05} & \metriccell{tablepoor!100!tablemid}{67.35\%} \\
SewingLDM~\cite{liu2025multimodal} & \metriccell{tablepoor!100!tablemid}{139.88} & \metriccell{tablepoor!20!tablemid}{30.17} & \metriccell{tablegood!20!tablemid}{0.730} & \metriccell{tablegood!100!tablemid}{\textbf{0.041}} & \metriccell{tablegood!100!tablemid}{\textbf{1.68}} & \metriccell{tablegood!33!tablemid}{95.31\%} & \metriccell{tablepoor!100!tablemid}{137.81} & \metriccell{tablepoor!60!tablemid}{24.59} & \metriccell{tablepoor!20!tablemid}{0.709} & \metriccell{tablegood!100!tablemid}{\textbf{0.578}} & \metriccell{tablegood!60!tablemid}{\underline{13.50}} & \metriccell{tablepoor!33!tablemid}{73.47\%} \\
Design2Cloth~\cite{zheng2024design2cloth} & \metriccell{tablepoor!20!tablemid}{104.48} & \metriccell{tablepoor!100!tablemid}{9.72} & \metriccell{tablepoor!60!tablemid}{0.708} & \metriccell{tablepoor!60!tablemid}{7.387} & \metriccell{tablepoor!60!tablemid}{27.61} & \metriccell{tablegood!100!tablemid}{\textbf{100.00\%}} & \metriccell{tablepoor!20!tablemid}{64.87} & \metriccell{tablepoor!100!tablemid}{12.15} & \metriccell{tablegood!20!tablemid}{0.755} & \metriccell{tablepoor!60!tablemid}{5.256} & \metriccell{tablegood!20!tablemid}{25.02} & \metriccell{tablegood!100!tablemid}{\textbf{100.00\%}} \\
Garment3DGen~\cite{sarafianos2025garment3dgen} & \metriccell{tablegood!60!tablemid}{\underline{20.19}} & \metriccell{tablegood!60!tablemid}{\underline{39.40}} & \metriccell{tablegood!60!tablemid}{\underline{0.770}} & \metriccell{tablepoor!100!tablemid}{31.007} & \metriccell{tablepoor!100!tablemid}{59.94} & \metriccell{tablegood!100!tablemid}{\textbf{100.00\%}} & \metriccell{tablegood!60!tablemid}{\underline{21.64}} & \metriccell{tablegood!60!tablemid}{\underline{39.65}} & \metriccell{tablegood!60!tablemid}{\underline{0.771}} & \metriccell{tablepoor!100!tablemid}{27.736} & \metriccell{tablepoor!100!tablemid}{44.84} & \metriccell{tablegood!100!tablemid}{\textbf{100.00\%}} \\
Ours & \metriccell{tablegood!100!tablemid}{\textbf{6.36}} & \metriccell{tablegood!100!tablemid}{\textbf{83.68}} & \metriccell{tablegood!100!tablemid}{\textbf{0.920}} & \metriccell{tablegood!60!tablemid}{\underline{0.080}} & \metriccell{tablegood!60!tablemid}{\underline{2.58}} & \metriccell{tablegood!100!tablemid}{\textbf{100.00\%}} & \metriccell{tablegood!100!tablemid}{\textbf{16.14}} & \metriccell{tablegood!100!tablemid}{\textbf{68.98}} & \metriccell{tablegood!100!tablemid}{\textbf{0.867}} & \metriccell{tablegood!20!tablemid}{0.856} & \metriccell{tablegood!100!tablemid}{\textbf{12.83}} & \metriccell{tablegood!100!tablemid}{\textbf{100.00\%}} \\
\bottomrule
\end{tabular*}
\end{table*}

\vspace{-1em}
\paragraph{Metrics and protocol.}
\label{sec:metrics}
We measure surface fidelity with non-squared symmetric Chamfer distance (CD), F-score at 10\,mm (F@10), and absolute normal consistency (Normal).
Body compatibility is assessed by penetration rate (Pen.), per-mesh 95th-percentile penetration depth (P95), and clearance error (Clr.), which measures deviation from ground-truth garment--body spacing.
Appearance uses PSNR, SSIM, and LPIPS.
Success (Succ.) reports the percentage of eligible cases with evaluable outputs, over which quality metrics are averaged.
Detailed metric definitions, surface sampling procedures, and result aggregation protocols are provided in Supplementary Section~\ref{sec:supp-evaluation}.

\subsection{Body-Aligned Garment Reconstruction}
\label{sec:reconstruction}
Matched-body reconstruction tests whether BodyWeave recovers garment geometry and placement when given the body associated with the reference garment.
Table~\ref{tab:reconstruction} compares this capability on the GarmentWeave3D evaluation subset and on out-of-domain Artist-Created Garments.
The latter tests reconstruction of complex silhouettes and fine structural details beyond the training domain.
BodyWeave achieves the lowest mean CD and highest F@10 and normal consistency on both collections.
On Artist-Created Garments, it reduces mean CD by \textbf{25.4\%} relative to Garment3DGen while maintaining 0.86\% penetration.

\Cref{fig:reconstruction} illustrates how design fidelity and body compatibility differ across the compared methods.
In these examples, simulation-based methods dress the body but often change the reference silhouette, sleeve structure, or garment length. Zoom for details of the geometry comparision.
Garment3DGen recovers aspects of the reference shape, but body surfaces protrude through the garment around the shoulders and torso.
BodyWeave more closely reproduces the reference silhouettes and garment components while fitting them to the body, including the sleeve details and separate trouser legs in the lower rows.
The normalized-shape evaluation in Supplementary Section~\ref{sec:shape} separately measures geometric fidelity after removing global placement.

\subsection{Adaptation to a New Target Body}
\label{sec:body-response}
We test adaptation to a new wearer by retaining source garment images and comparing predictions with the same design simulated on a different target body.
Table~\ref{tab:adaptation} reports the lowest target CD and clearance error for BodyWeave across 16 directed tasks.
\Cref{fig:adaptation} shows that BodyWeave preserves skirt pleats and the fitted long-dress silhouette while matching target sleeve lengths and hem positions.
The displayed baselines change these proportions or leave visible body intersections.
Further evaluation details and results are provided in Supplementary Section~\ref{sec:supp-body}.

\begin{table}[!tbp]
\caption{Cross-body adaptation on target-body tasks. \textbf{Bold} and \underline{underlined} values mark the top two results.}
\label{tab:adaptation}
\centering\footnotesize\setlength{\tabcolsep}{2pt}\renewcommand{\arraystretch}{1.15}
\begin{tabular*}{\columnwidth}{@{}>{\raggedright\arraybackslash}p{\dimexpr0.32\columnwidth-\tabcolsep\relax}*{4}{>{\raggedleft\arraybackslash}p{\dimexpr0.17\columnwidth-2\tabcolsep\relax}}@{\hspace{\tabcolsep}}}
\toprule
\textbf{Method} & \textbf{CD$\downarrow$} & \textbf{Pen.$\downarrow$} & \textbf{Clr.$\downarrow$} & {\scriptsize\textbf{Succ.\%$\uparrow$}} \\
\midrule
Garment Particles~\cite{nakayama2026garment} & \metriccell{tablepoor!100!tablemid}{91.13} & \metriccell{tablegood!20!tablemid}{0.243} & \metriccell{tablepoor!20!tablemid}{21.63} & \metriccell{tablegood!100!tablemid}{\textbf{100.00\%}} \\
AIpparel~\cite{nakayama2025aipparel} & \metriccell{tablegood!60!tablemid}{\underline{58.21}} & \metriccell{tablepoor!20!tablemid}{0.304} & \metriccell{tablegood!20!tablemid}{18.70} & \metriccell{tablepoor!100!tablemid}{81.25\%} \\
SewingLDM~\cite{liu2025multimodal} & \metriccell{tablepoor!20!tablemid}{78.66} & \metriccell{tablegood!100!tablemid}{\textbf{$3\!\times\!10^{-4}$}} & \metriccell{tablegood!60!tablemid}{\underline{15.10}} & \metriccell{tablegood!100!tablemid}{\textbf{100.00\%}} \\
Design2Cloth~\cite{zheng2024design2cloth} & \metriccell{tablepoor!60!tablemid}{90.18} & \metriccell{tablepoor!60!tablemid}{7.045} & \metriccell{tablepoor!60!tablemid}{38.77} & \metriccell{tablegood!100!tablemid}{\textbf{100.00\%}} \\
Garment3DGen~\cite{sarafianos2025garment3dgen} & \metriccell{tablegood!20!tablemid}{74.78} & \metriccell{tablepoor!100!tablemid}{12.994} & \metriccell{tablepoor!100!tablemid}{45.11} & \metriccell{tablegood!100!tablemid}{\textbf{100.00\%}} \\
Ours & \metriccell{tablegood!100!tablemid}{\textbf{8.78}} & \metriccell{tablegood!60!tablemid}{\underline{0.204}} & \metriccell{tablegood!100!tablemid}{\textbf{4.85}} & \metriccell{tablegood!100!tablemid}{\textbf{100.00\%}} \\
\bottomrule
\end{tabular*}
\end{table}

\begin{table}[!tbp]
\caption{Appearance synthesis on shared geometry. \textbf{Bold} and \underline{underlined} values mark the top two results.}
\label{tab:appearance}
\centering\footnotesize\setlength{\tabcolsep}{2pt}\renewcommand{\arraystretch}{1.15}
\begin{tabular*}{\columnwidth}{@{}>{\raggedright\arraybackslash}p{\dimexpr0.43\columnwidth-\tabcolsep\relax}*{3}{>{\raggedleft\arraybackslash}p{\dimexpr0.19\columnwidth-2\tabcolsep\relax}}@{\hspace{\tabcolsep}}}
\toprule
\textbf{Method} & \textbf{PSNR$\uparrow$} & \textbf{SSIM$\uparrow$} & \textbf{LPIPS$\downarrow$} \\
\midrule
\multicolumn{4}{l}{Input view (front)} \\
\hspace{0.6em}FabricDiffusion~\cite{zhang2024fabricdiffusion} & \metriccell{tablepoor!100!tablemid}{13.62} & \metriccell{tablepoor!100!tablemid}{0.495} & \metriccell{tablepoor!100!tablemid}{0.324} \\
\hspace{0.6em}TRELLIS.2~\cite{xiang2026native} & \metriccell{tablegood!0!tablemid}{\underline{15.35}} & \metriccell{tablegood!0!tablemid}{\underline{0.658}} & \metriccell{tablegood!0!tablemid}{\underline{0.189}} \\
\hspace{0.6em}Ours & \metriccell{tablegood!100!tablemid}{\textbf{23.03}} & \metriccell{tablegood!100!tablemid}{\textbf{0.802}} & \metriccell{tablegood!100!tablemid}{\textbf{0.112}} \\
\midrule
\multicolumn{4}{l}{Held-out views (left/right)} \\
\hspace{0.6em}FabricDiffusion~\cite{zhang2024fabricdiffusion} & \metriccell{tablepoor!100!tablemid}{14.06} & \metriccell{tablepoor!100!tablemid}{0.518} & \metriccell{tablepoor!100!tablemid}{0.360} \\
\hspace{0.6em}TRELLIS.2~\cite{xiang2026native} & \metriccell{tablegood!0!tablemid}{\underline{15.25}} & \metriccell{tablegood!0!tablemid}{\underline{0.660}} & \metriccell{tablegood!0!tablemid}{\underline{0.223}} \\
\hspace{0.6em}Ours & \metriccell{tablegood!100!tablemid}{\textbf{22.11}} & \metriccell{tablegood!100!tablemid}{\textbf{0.783}} & \metriccell{tablegood!100!tablemid}{\textbf{0.159}} \\
\bottomrule
\end{tabular*}
\end{table}

\subsection{Appearance on Fixed Geometry}
\label{sec:appearance-eval}
\begin{figure}[!tbp]
\centering
\includegraphics[width=\columnwidth]{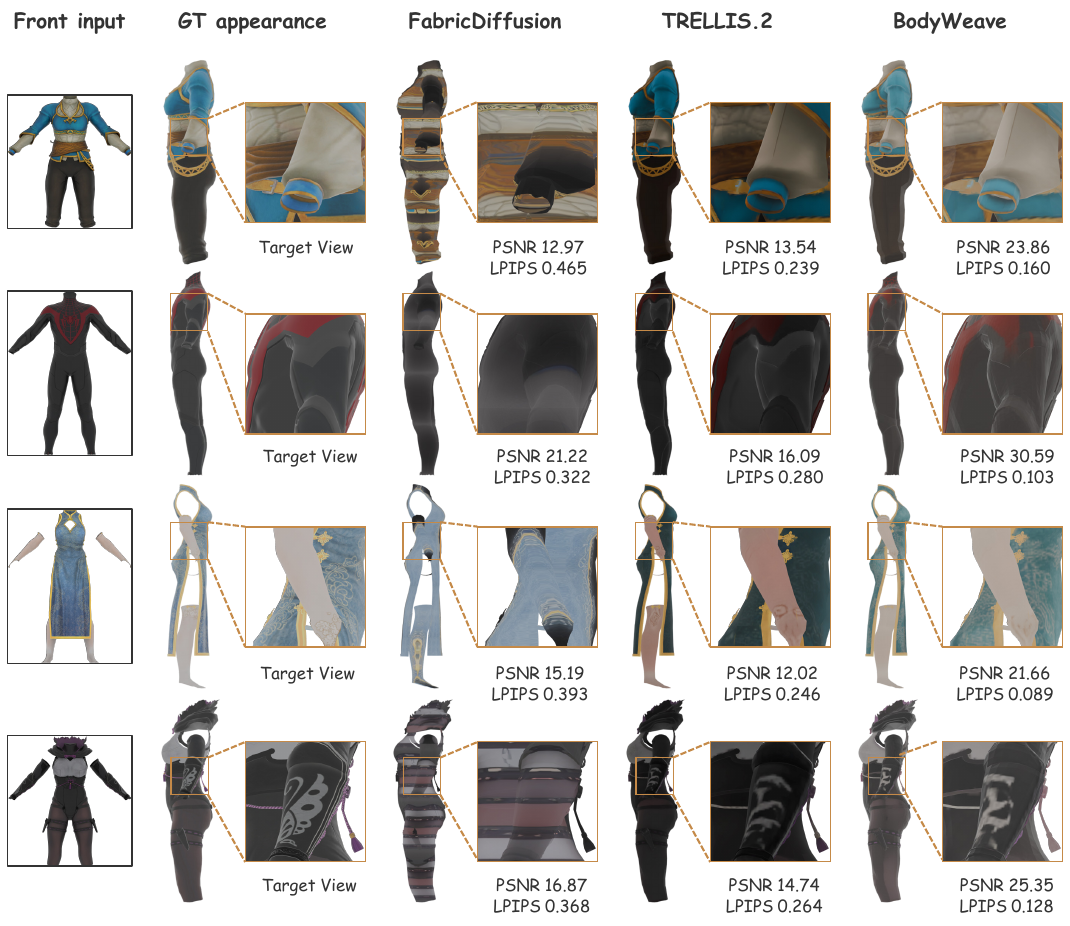}
\caption{Appearance synthesis on shared geometry. Each row shows the front input, held-out target view, and detail crops for the reference and generated appearances. Zoom for details.}
\label{fig:appearance}
\end{figure}

We fix the ground-truth geometry and rendering conditions to isolate appearance synthesis from shape reconstruction.
On Artist-Created Garments, all methods receive only the front view, with left and right views held out to evaluate appearance beyond the input.
Table~\ref{tab:appearance} shows that BodyWeave achieves the best PSNR, SSIM, and LPIPS in both input and held-out views.
Its held-out LPIPS is \textbf{28.9\% lower} than TRELLIS.2's.
\Cref{fig:appearance} shows misplaced bands and repeated patterns from FabricDiffusion, and altered reference colors and missing decorative details from TRELLIS.2.
BodyWeave more closely preserves the sleeve trim, floral motifs, and printed sleeve pattern in the held-out views.
The complete evaluation protocol, metrics, and material-channel visualizations are provided in Supplementary Section~\ref{sec:supp-appearance}.

\begin{table}[!tbp]
\caption{Body-conditioning and penetration-loss ablations. \textbf{Bold} and \underline{underlined} values mark the top two results.}
\label{tab:ablation}
\centering\footnotesize\setlength{\tabcolsep}{0pt}\renewcommand{\arraystretch}{1.15}
\begin{tabular*}{\columnwidth}{@{}>{\raggedright\arraybackslash}p{0.28\columnwidth}*{4}{>{\raggedleft\arraybackslash}p{0.18\columnwidth}}@{}}
\toprule
\textbf{Setting} & \textbf{R-CD$\downarrow$} & \textbf{T-CD$\downarrow$} & \textbf{Pen.$\downarrow$} & \textbf{Clr.$\downarrow$} \\
\midrule
GBA off & 279.20 & 275.50 & \underline{0.311} & 194.10 \\
LBA off & \underline{7.11} & \underline{9.94} & 9.338 & \underline{6.11} \\
Full & \textbf{6.36} & \textbf{8.78} & \textbf{0.204} & \textbf{4.85} \\
\midrule
\textbf{Setting} & \textbf{Mean$\downarrow$} & \textbf{Max.$\downarrow$} & \textbf{Pen.$\downarrow$} & \textbf{CD$\downarrow$} \\
w/o pen. loss & \underline{0.0156} & \underline{6.60} & \underline{0.229} & \underline{8.88} \\
w/ pen. loss & \textbf{0.0111} & \textbf{4.54} & \textbf{0.186} & \textbf{8.80} \\
\bottomrule
\end{tabular*}
\end{table}

\subsection{Ablation and Design Analysis}
\label{sec:ablation}
\begin{figure}[!tbp]
\centering
\includegraphics[width=\columnwidth]{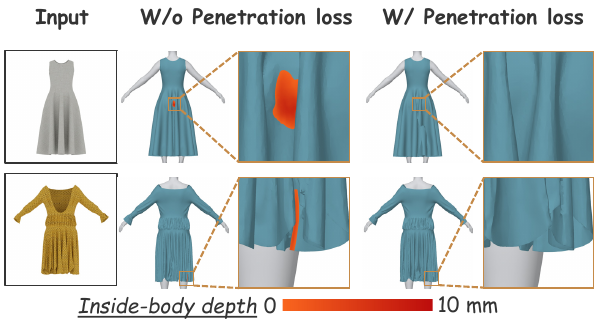}
\caption{Garment--body intersections with and without surface penetration supervision. Color indicates depth scale.}
\label{fig:penetration}
\end{figure}

\begin{figure}[!tbp]
\centering
\includegraphics[width=\columnwidth]{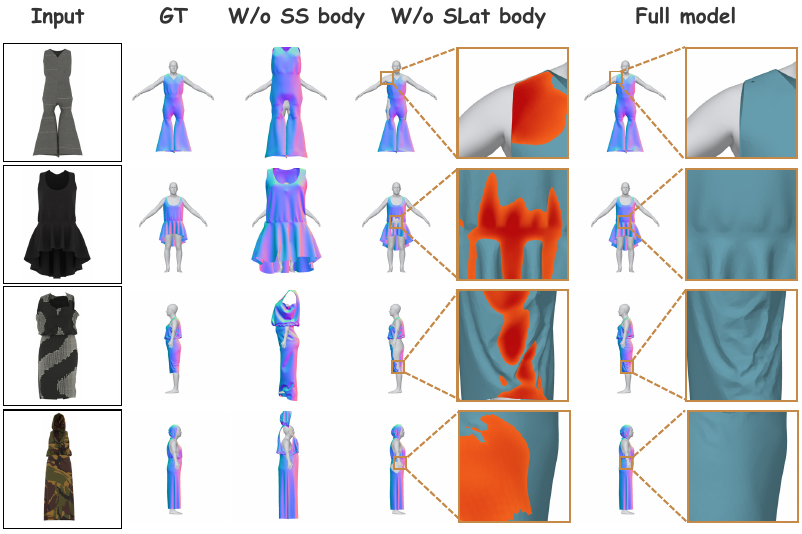}
\caption{Visual ablation of body conditioning. Each row shows the input garment, ground truth, and outputs without global SS conditioning, without local SLat conditioning, and with the full model. Insets magnify the corresponding regions.}
\label{fig:body-conditioning}
\end{figure}

\paragraph{Hierarchical body conditioning.}
To examine the roles of global and local body context, we mask GBA or LBA at inference in the same trained checkpoint.
The upper panel of Table~\ref{tab:ablation} shows large placement errors without GBA and increased penetration without LBA.
\Cref{fig:body-conditioning} makes this distinction visible: masking GBA produces garments with incorrect scale or position relative to the wearer, while masking LBA retains the overall placement but introduces local intersections around the shoulder, waist, and hip.
The low penetration rate without GBA does not imply correct fitting, as the large placement and clearance errors show.
The full model reduces these intersections while retaining garment folds, supporting complementary roles for global placement and local body compatibility.
\vspace{-1em}
\paragraph{Penetration supervision.}
To assess the effect of explicit surface supervision alongside latent flow matching, we compare matched-step fine-tuning with and without Surface Penetration Loss.
The lower panel of Table~\ref{tab:ablation} reports a \textbf{31.1\% reduction} in mean per-mesh sampled maximum penetration depth, with nearly unchanged target CD.
\Cref{fig:penetration} shows that surface supervision reduces highlighted intersections while largely preserving dress silhouettes and folds.
Additional contact-severity metrics and examples appear in Supplementary Table~\ref{tab:supp-penetration-severity} and Figure~\ref{fig:supp-penetration}.

\section{Conclusion}
\label{sec:conclusion}
We present \method, which uses the 3D target body as a native condition for image-guided garment generation.
Its key innovation combines hierarchical body conditioning for global placement and local refinement with decoded-surface penetration supervision.
We also introduce \dataset, integrating detailed garment assets, same-design cross-body pairs, and fabric materials.
Experiments demonstrate faithful reconstruction, body adaptation, and appearance synthesis, including strong reconstruction of complex silhouettes and fine details on out-of-domain Artist-Created Garments.
\vspace{-1em}
\paragraph{Limitation.}
The framework assumes orthographic views and static garments, leaving unconstrained images and temporally coherent garment motion for future work.

\clearpage
\appendix
\onecolumn
\begingroup
\raggedbottom
\captionsetup[table]{justification=raggedright,singlelinecheck=false}
\setlength{\textfloatsep}{12pt plus 2pt minus 2pt}
\begin{center}
{\Large\bfseries \thetitle\par}
\vspace{0.5em}
{\large\bfseries Supplementary Material\par}
\end{center}
\medskip
This supplement provides extended related work (\cref{sec:supp-background}), dataset construction (\cref{sec:dataset-construction}), additional experiments and ablations (\cref{sec:supp-experiments,sec:supp-ablation}), evaluation protocols (\cref{sec:supp-protocols}), training and inference details (\cref{sec:method-details}), and qualitative comparisons (\cref{sec:supp-qualitative}).

\setcounter{table}{0}
\tableformat
\setcounter{figure}{0}
\renewcommand{\thetable}{S\arabic{table}}
\renewcommand{\thefigure}{S\arabic{figure}}
\renewcommand{\theHtable}{supp.\arabic{table}}
\renewcommand{\theHfigure}{supp.\arabic{figure}}
\section{Background and Extended Related Work}
\label{sec:supp-background}

\subsection{Structured Garment Representations and Data}
Sewing patterns represent garment construction through panels and stitching relationships.
This structure connects design parameters to garment assembly and simulation on a target body.
Korosteleva and Lee~\cite{korosteleva2021generating} provide paired 3D garments and patterns; GarmentCode offers programmable patterns~\cite{korosteleva2023garmentcode}; and GarmentCodeData supplies made-to-measure garments across body shapes~\cite{korosteleva2024garmentcodedata}.
\dataset combines public garment--body observations with curated high-detail assets and fabric materials, retaining spatial alignment and same-design cross-body correspondences (\cref{sec:dataset-construction}).

GarmentDiffusion generates vectorized patterns with a multimodal diffusion transformer~\cite{li2025garmentdiffusion}.
SwiftTailor reconstructs surfaces through pattern-conditioned geometry images and mesh assembly~\cite{pham2026swifttailor}, while Stitched Embeddings links pattern inference and 3D reconstruction through shared latents and a BoxMesh intermediate representation~\cite{sanchietti2026stitched}.
ReWeaver recovers panels and seam connectivity in 2D and 3D from sparse views~\cite{li2026reweaver}.
These methods connect garment construction to surface recovery; \method directly learns body-dependent surface geometry from aligned observations.

\subsection{Visual Reconstruction and Generative Priors}
Visual reconstruction uses image evidence to recover garment silhouettes and geometric detail.
ClothWild uses a learned garment representation for clothed-human reconstruction~\cite{moon20223d}; GarVerseLOD uses artist-authored data and hierarchical geometric detail~\cite{luo2024garverselod}.
GarmentCrafter improves view consistency through depth prediction, warping, and completion~\cite{wang2026garmentcrafter}, while MagicGarment benchmarks standalone reconstruction and adapts an image-to-3D backbone to garments~\cite{chen2026magicgarment}.

Wonder3D and InstantMesh reconstruct geometry through multi-view generation~\cite{long2024wonder3d,xu2024instantmesh}; CLAY and TRELLIS learn latent 3D generative priors~\cite{zhang2024clay,xiang2025structured}.
These priors provide flexible surface representations for varied designs. Adapting an image-defined garment to a separately specified wearer additionally requires the target body's spatial context.
\method extends TRELLIS.2's geometry and PBR representation~\cite{xiang2026native} with target-body conditioning, allowing fixed reference images to guide garment generation for different wearers.

\subsection{Body Conditioning in Draping and Generation}
Body-conditioned garment models relate clothing geometry to the wearer's shape and pose.
TailorNet predicts clothing deformation from pose, shape, and garment style~\cite{patel2020tailornet}, while SMPLicit represents clothing with a body-conditioned implicit model~\cite{corona2021smplicit}.
DIG learns differentiable draping for implicit garments~\cite{li2022dig}, while DrapeNet combines a garment prior with self-supervised draping~\cite{de2023drapenet}.
ISP combines implicit panel boundaries with 2D-to-3D mappings for layered draping~\cite{li2023isp}.

Generative pipelines incorporate body geometry at different stages.
Dress-1-to-3 refines patterns through differentiable simulation guided by generated views~\cite{li2025dress}; BAG conditions image generation on body projections before 3D alignment and proxy-based collision handling~\cite{luo2026bag}; and Tailor constrains text-guided template deformation with body geometry~\cite{sun2026text}.
In \method, GBA and LBA inject body features directly into sparse-support and surface-latent generation, while decoded-surface supervision constrains intersections during training.
This places global body context and local surface information within the pretrained generator, supporting joint learning of garment shape and fit.

\subsection{Fabric Appearance and Material Generation}
FabricDiffusion transfers fabric appearance from images to 3D garments~\cite{zhang2024fabricdiffusion}, while OmniFabric refines garment textures in a canonical sewing-pattern UV domain~\cite{huang2026omnifabric}.
\method fine-tunes TRELLIS.2's geometry-conditioned appearance model~\cite{xiang2026native} with curated garment materials to predict base color, metallic, roughness, and alpha.
The generated geometry provides spatial support for these attributes, while reference images specify the garment's appearance.
Shared-geometry comparisons assess appearance fidelity in input and held-out views (\cref{sec:supp-appearance}).

\section{Dataset Construction}
\label{sec:dataset-construction}

\subsection{Geometry Sources and Body Pairing}
\dataset combines three complementary sources to cover body variation, garment designs, and fine structural detail.
GarmentCodeData (v2)~\cite{korosteleva2024garmentcodedata} contributes 229,448 observations of the same designs simulated on neutral and sampled bodies, supplying all 114,724 same-design cross-body pairs.
The Sewing-Pattern Dataset of Korosteleva and Lee~\cite{korosteleva2021generating} broadens design coverage with 33,419 observations simulated on a T-pose body.
Our curated GarmentWeave3D-Detail collection contributes 30,898 observations settled on an A-pose body, including assets with dense sleeve gathers, hood constructions, and layered shoulder details.
Together, these sources provide 293,765 garment--body observations before filtering. Each observation contains a garment and its associated body; the 114,724 cross-body pairs link two observations of the same design.

All observations retain their simulated garment--body relationships in a shared coordinate frame and physical scale.
Matched reconstruction pairs reference images with their associated body and garment geometry.
Cross-body supervision holds the source garment views fixed while replacing the target body and corresponding garment geometry with the other member of a same-design pair.
These pairs supervise body-dependent geometric changes under a fixed visual design.

\subsection{Dataset Splits}
\label{sec:dataset-splits}
The dataset contains 287,765 observations after excluding 6,000 Sewing-Pattern observations with garment--body penetration.
Training, validation, and test partitions contain 261,277, 13,254, and 13,234 observations, respectively (\cref{tab:dataset-splits}).
GarmentCodeData retains its official design-level partition, with both body versions of each design assigned together.
For the other sources, the final validation and test subsets are selected from deterministic initial partitions, keeping checksum-linked geometry components together.
Assigned materials and rendered views follow their corresponding geometry partition.

\begin{center}
\begin{minipage}{\linewidth}
\captionsetup{type=table}
\caption{Garment--body observations by source and partition. The chart shows each source's share of the 287,765 observations.}
\label{tab:dataset-splits}
\begin{minipage}[c]{.49\linewidth}
\centering\footnotesize
\setlength{\tabcolsep}{2pt}
\begin{tabular*}{\linewidth}{@{\extracolsep{\fill}}lrrr@{}}
\toprule
Source & Train & Validation & Test \\
\midrule
GarmentCodeData (v2) & 204,458 & 12,500 & 12,490 \\
Sewing-Pattern Dataset & 26,699 & 363 & 357 \\
GarmentWeave3D-Detail & 30,120 & 391 & 387 \\
\midrule
Total & 261,277 & 13,254 & 13,234 \\
\bottomrule
\end{tabular*}
\end{minipage}\hfill
\begin{minipage}[c]{.49\linewidth}
\centering
\definecolor{splitgcd}{RGB}{71,105,145}
\definecolor{splitsewing}{RGB}{211,158,96}
\definecolor{splitdetail}{RGB}{100,157,150}
\definecolor{splittext}{RGB}{53,63,74}
\begin{tikzpicture}[font=\scriptsize]
\fill[splitgcd] (0,0) -- (126:1.13cm) arc (126:413.04422:1.13cm) -- cycle;
\fill[splitsewing] (0,0) -- (413.04422:1.13cm) arc (413.04422:447.34596:1.13cm) -- cycle;
\fill[splitdetail] (0,0) -- (447.34596:1.13cm) arc (447.34596:486:1.13cm) -- cycle;
\fill[white] (0,0) circle (.76cm);
\foreach \angle in {126,413.04422,447.34596}
  \draw[white,line width=1.2pt] (\angle:.75cm) -- (\angle:1.14cm);
\node[text=splittext,font=\small\bfseries] at (0,.12) {287,765};
\node[text=splittext!75,font=\tiny] at (0,-.22) {observations};
\fill[splitgcd,rounded corners=.5pt] (1.43,.76) rectangle (1.58,.91);
\node[anchor=west,text=splittext,inner sep=0pt] at (1.72,.85) {GarmentCodeData (v2)};
\node[anchor=west,text=splitgcd,inner sep=0pt,font=\scriptsize\bfseries] at (1.72,.54) {79.7\%};
\fill[splitsewing,rounded corners=.5pt] (1.43,.06) rectangle (1.58,.21);
\node[anchor=west,text=splittext,inner sep=0pt] at (1.72,.15) {Sewing-Pattern Dataset};
\node[anchor=west,text=splittext!75,inner sep=0pt,font=\scriptsize\bfseries] at (1.72,-.16) {9.5\%};
\fill[splitdetail,rounded corners=.5pt] (1.43,-.64) rectangle (1.58,-.49);
\node[anchor=west,text=splittext,inner sep=0pt] at (1.72,-.55) {GarmentWeave3D-Detail};
\node[anchor=west,text=splittext!75,inner sep=0pt,font=\scriptsize\bfseries] at (1.72,-.86) {10.7\%};
\end{tikzpicture}
\end{minipage}
\end{minipage}
\end{center}

\subsection{Fabric Material Curation}
\label{sec:material-curation}
We select 2,495 apparel-suitable fabric materials from 5,114 artist-authored candidates.
Each retained material must load correctly, render consistently, and provide valid base-color, metallic, roughness, and alpha attributes.
We exclude emissive, purely metallic, transparent, and alpha-textured materials, as well as assets with invalid attributes or rendering failures.
This produces opaque appearance targets with consistent silhouette supervision.
Both members of each same-design cross-body pair share the same material assignment, preserving appearance identity as geometry changes with the wearer.

\subsection{Rendering and Image Conditions}
We render each textured garment from orthographic front, left, and right views at a common camera scale across views.
These views match the multiview design references targeted by our task, with side views revealing silhouettes and surface regions absent from the frontal view.
Rendering the garment alone makes the images describe its design and appearance, while the separately supplied body specifies the target wearer.
The renders provide image conditions and, together with assigned PBR attributes, appearance supervision.
All stages use the same garment-only renderings, resized to $512^2$ for SS/LR and retained at $1024^2$ for HR/PBR; all 3D supervision remains in the body coordinate frame.

\section{Experiments}
\label{sec:supp-experiments}
The following experiments examine target-specific body adaptation, garment shape fidelity, and appearance synthesis.
Body-condition interventions separate the influence of the target body from the fixed image reference, while normalized-shape and shared-geometry evaluations distinguish geometric and appearance fidelity from global placement.
Evaluation protocols are detailed in \cref{sec:supp-protocols}.

\subsection{Target-Specific Body Adaptation}
\label{sec:supp-body}
We evaluate adaptation by keeping the source garment images fixed and supplying a different target body, with the same design simulated on that body as the reference.
The cross-body cohort is drawn from GarmentCodeData (v2)~\cite{korosteleva2024garmentcodedata} and contains eight independent designs, each evaluated in both directions for a total of 16 tasks.
BodyWeave achieves the lowest target CD, clearance error, and inflation among the compared methods (\cref{tab:s2}), recovering target-body geometry while preserving garment--body spacing.

Body-condition interventions reveal whether this adaptation responds to the supplied wearer.
We hold the source images, model, and random seed fixed and supply the source body, a mismatched body, or the correct target body; geometric errors and penetration are always measured against the correct target (\cref{tab:s2b}).
Replacing the source body with the correct target reduces CD from 18.45 to 8.78\,mm and penetration from 22.98\% to 0.20\%.
The correct target also yields lower geometric error and penetration than the mismatched body, demonstrating target-specific control of garment geometry and fit.
\begin{center}
\begin{minipage}{\linewidth}
\captionsetup{type=table}
\footnotesize\setlength{\tabcolsep}{4pt}
\caption{Cross-body garment adaptation on 16 directed tasks from eight designs. CD, clearance, and inflation are in mm; F@10, T-dir., penetration, and success are percentages. Target-change direction (T-dir.) uses the source GT garment as reference.}\label{tab:s2}
\centering
\begin{tabular*}{\linewidth}{@{}>{\raggedright\arraybackslash}p{\dimexpr0.28000000\linewidth-\tabcolsep\relax}*{7}{>{\raggedleft\arraybackslash}p{\dimexpr0.10285714\linewidth-2\tabcolsep\relax}}@{\hspace{\tabcolsep}}}
\toprule
\textbf{Method} & \textbf{CD$\downarrow$} & \textbf{F@10$\uparrow$} & \textbf{T-dir.$\uparrow$} & \textbf{Pen.$\downarrow$} & \textbf{Clearance$\downarrow$} & \textbf{Inflation$\downarrow$} & \textbf{Succ.\%$\uparrow$} \\
\midrule
Garment Particles~\cite{nakayama2026garment} & \metriccell{tablepoor!100!tablemid}{91.13} & \metriccell{tablepoor!20!tablemid}{26.21} & \metriccell{tablegood!20!tablemid}{52.08} & \metriccell{tablegood!20!tablemid}{0.243} & \metriccell{tablepoor!20!tablemid}{21.63} & \metriccell{tablepoor!20!tablemid}{17.56} & \metriccell{tablegood!100!tablemid}{\textbf{100.00\%}} \\
AIpparel~\cite{nakayama2025aipparel} & \metriccell{tablegood!60!tablemid}{\underline{58.21}} & \metriccell{tablegood!60!tablemid}{\underline{37.66}} & \metriccell{tablepoor!100!tablemid}{44.87} & \metriccell{tablepoor!20!tablemid}{0.304} & \metriccell{tablegood!20!tablemid}{18.70} & \metriccell{tablegood!20!tablemid}{14.63} & \metriccell{tablepoor!100!tablemid}{81.25\%} \\
SewingLDM~\cite{liu2025multimodal} & \metriccell{tablepoor!20!tablemid}{78.66} & \metriccell{tablegood!20!tablemid}{35.70} & \metriccell{tablegood!60!tablemid}{\underline{54.17}} & \metriccell{tablegood!100!tablemid}{\textbf{$3\!\times\!10^{-4}$}} & \metriccell{tablegood!60!tablemid}{\underline{15.10}} & \metriccell{tablegood!60!tablemid}{\underline{6.99}} & \metriccell{tablegood!100!tablemid}{\textbf{100.00\%}} \\
Design2Cloth~\cite{zheng2024design2cloth} & \metriccell{tablepoor!60!tablemid}{90.18} & \metriccell{tablepoor!100!tablemid}{11.87} & \metriccell{tablepoor!60!tablemid}{47.92} & \metriccell{tablepoor!60!tablemid}{7.045} & \metriccell{tablepoor!60!tablemid}{38.77} & \metriccell{tablepoor!60!tablemid}{35.92} & \metriccell{tablegood!100!tablemid}{\textbf{100.00\%}} \\
Garment3DGen~\cite{sarafianos2025garment3dgen} & \metriccell{tablegood!20!tablemid}{74.78} & \metriccell{tablepoor!60!tablemid}{12.00} & \metriccell{tablepoor!20!tablemid}{48.96} & \metriccell{tablepoor!100!tablemid}{12.994} & \metriccell{tablepoor!100!tablemid}{45.11} & \metriccell{tablepoor!100!tablemid}{38.94} & \metriccell{tablegood!100!tablemid}{\textbf{100.00\%}} \\
Ours & \metriccell{tablegood!100!tablemid}{\textbf{8.78}} & \metriccell{tablegood!100!tablemid}{\textbf{75.55}} & \metriccell{tablegood!100!tablemid}{\textbf{78.13}} & \metriccell{tablegood!60!tablemid}{\underline{0.204}} & \metriccell{tablegood!100!tablemid}{\textbf{4.85}} & \metriccell{tablegood!100!tablemid}{\textbf{2.64}} & \metriccell{tablegood!100!tablemid}{\textbf{100.00\%}} \\
\bottomrule
\end{tabular*}
\end{minipage}
\end{center}
\begin{center}
\begin{minipage}{\linewidth}
\captionsetup{type=table}
\footnotesize\setlength{\tabcolsep}{4pt}
\caption{Body-condition interventions with fixed images, model, and seed. All outputs are evaluated against the correct target. Distance metrics are in mm; T-dir., penetration, and success are percentages.}\label{tab:s2b}
\centering
\begin{tabular*}{\linewidth}{@{\extracolsep{\fill}}lrrrrrrr@{}}
\toprule
\textbf{Condition} & \textbf{CD$\downarrow$} & \textbf{DINO$\uparrow$} & \textbf{T-dir.$\uparrow$} & \textbf{Pen.$\downarrow$} & \textbf{Clearance$\downarrow$} & \textbf{Inflation$\downarrow$} & \textbf{Succ.\%$\uparrow$} \\
\midrule
Ours (source) & 18.45 & \underline{0.450} & 39.58 & \underline{22.981} & 14.71 & 7.57 & \textbf{100.00\%} \\
Ours (wrong) & \underline{16.43} & \textbf{0.452} & \underline{63.54} & 23.216 & \underline{12.43} & \underline{4.17} & \textbf{100.00\%} \\
Ours (correct) & \textbf{8.78} & 0.443 & \textbf{78.13} & \textbf{0.204} & \textbf{4.85} & \textbf{2.64} & \textbf{100.00\%} \\
\bottomrule
\end{tabular*}
\end{minipage}
\end{center}
\Cref{fig:supp-adaptation-comparison} illustrates the resulting garment shapes and target-body fit across three designs.
\begin{figure}[!tbp]
\centering
\includegraphics[width=\textwidth]{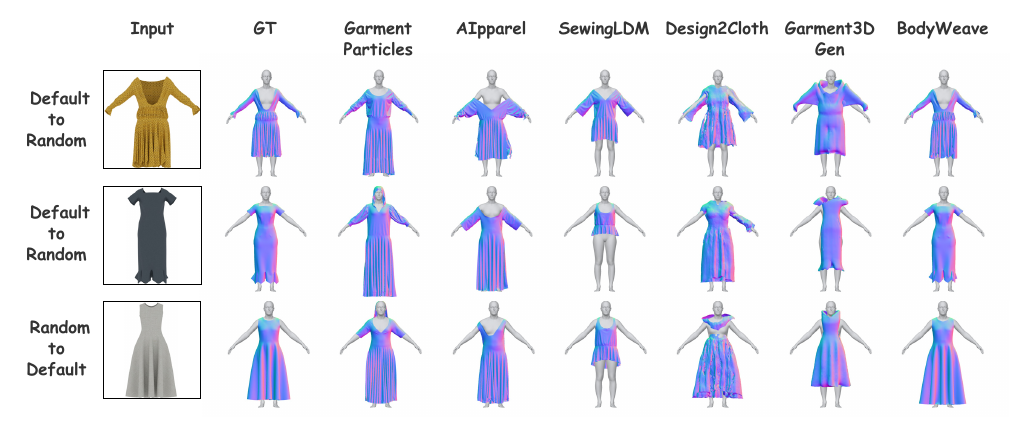}
\caption{Cross-body adaptation examples for three garment designs, shown as normal renderings.}
\label{fig:supp-adaptation-comparison}
\end{figure}

\subsection{Normalized Shape Fidelity}
\label{sec:shape}
Normalized evaluation measures garment shape fidelity independently of global placement and scale on Artist-Created Garments.
After conversion to a common coordinate frame, each predicted and reference garment is independently AABB-centered and normalized by its maximum side length.
\Cref{tab:s3} compares complete systems with their native training, view counts, and body inputs; CD is dimensionless and the F-score threshold is 1\% of the normalized unit.
BodyWeave achieves a normalized CD of 0.0096 and F-score of 71.38\%, compared with 0.0113 and 60.72\% for TRELLIS.2.
The gains persist after removing translation and scale, indicating improved reconstruction of garment shape. \Cref{fig:supp-controls} illustrates the recovered silhouettes and surface details.

\subsection{Body Adaptation and Fixed-Geometry Placement}
Adapting a garment to a new wearer requires changes in shape as well as placement.
We compare BodyWeave with a control that repositions one fixed TRELLIS.2 mesh for the source and target bodies using a common scale and body-relative translation (\cref{sec:supp-protocol}).
The comparison uses eight designs with default-body source images. BodyWeave receives fixed F/L/R source views, while TRELLIS.2 receives the front view.
CD and F@10 average over source- and target-body reconstructions; response metrics measure the change between the two predictions for each design.

BodyWeave follows the GT response direction in 72.92\% of measurements, compared with 18.75\% for the fixed-geometry control (\cref{tab:s4}).
This comparison shows how body-conditioned generation captures garment changes under the evaluated placement protocol, complementing the within-model intervention in \cref{tab:s2b}.
\Cref{fig:supp-adaptation} illustrates the changes in geometry under fixed reference images, together with the generated appearance.

\begin{center}
\begin{minipage}[t]{.49\linewidth}
\vspace{0pt}
\captionsetup{type=table}
\small
\setlength{\tabcolsep}{3pt}
\parbox[t][40pt]{\linewidth}{\caption{Normalized shape fidelity on Artist-Created Garments. CD is dimensionless; F@1\% is a percentage.}\label{tab:s3}}
\par\smallskip\centering
\begin{tabular*}{\linewidth}{@{}>{\raggedright\arraybackslash}p{\dimexpr0.43000000\linewidth-\tabcolsep\relax}*{3}{>{\raggedleft\arraybackslash}p{\dimexpr0.19000000\linewidth-2\tabcolsep\relax}}@{\hspace{\tabcolsep}}}
\toprule
Method & CD$\downarrow$ & F@1\%$\uparrow$ & NC$\uparrow$ \\
\midrule
InstantMesh~\cite{xu2024instantmesh} & \metriccell{tablepoor!33!tablemid}{0.0199} & \metriccell{tablepoor!33!tablemid}{41.63} & \metriccell{tablepoor!33!tablemid}{0.791} \\
Garment3DGen~\cite{sarafianos2025garment3dgen} & \metriccell{tablepoor!100!tablemid}{0.0222} & \metriccell{tablepoor!100!tablemid}{37.25} & \metriccell{tablepoor!100!tablemid}{0.745} \\
TRELLIS.2~\cite{xiang2026native} & \metriccell{tablegood!33!tablemid}{\underline{0.0113}} & \metriccell{tablegood!33!tablemid}{\underline{60.72}} & \metriccell{tablegood!33!tablemid}{\underline{0.819}} \\
Ours & \metriccell{tablegood!100!tablemid}{\textbf{0.0096}} & \metriccell{tablegood!100!tablemid}{\textbf{71.38}} & \metriccell{tablegood!100!tablemid}{\textbf{0.864}} \\
\bottomrule
\end{tabular*}
\end{minipage}\hfill
\begin{minipage}[t]{.49\linewidth}
\vspace{0pt}
\captionsetup{type=table}
\small
\setlength{\tabcolsep}{3pt}
\parbox[t][40pt]{\linewidth}{\caption{Fixed-geometry placement control. Response direction compares source and target predictions.}\label{tab:s4}}
\par\smallskip\centering
\begin{tabular*}{\linewidth}{@{}>{\raggedright\arraybackslash}p{\dimexpr0.54000000\linewidth-\tabcolsep\relax}*{2}{>{\raggedleft\arraybackslash}p{\dimexpr0.23000000\linewidth-2\tabcolsep\relax}}@{\hspace{\tabcolsep}}}
\toprule
Metric & TRELLIS.2 & Ours \\
\midrule
CD (mm)$\downarrow$ & \metriccell{tablepoor!100!tablemid}{102.80} & \metriccell{tablegood!100!tablemid}{\textbf{7.01}} \\
F@10 (\%)$\uparrow$ & \metriccell{tablepoor!100!tablemid}{15.42} & \metriccell{tablegood!100!tablemid}{\textbf{81.86}} \\
Response error (mm)$\downarrow$ & \metriccell{tablepoor!100!tablemid}{20.41} & \metriccell{tablegood!100!tablemid}{\textbf{19.61}} \\
R-dir. (\%)$\uparrow$ & \metriccell{tablepoor!100!tablemid}{18.75} & \metriccell{tablegood!100!tablemid}{\textbf{72.92}} \\
\bottomrule
\end{tabular*}
\end{minipage}
\end{center}

\begin{center}
\begin{minipage}[t]{.49\linewidth}
\vspace{0pt}
\captionsetup{type=figure}
\centering
\includegraphics[width=\linewidth,height=.20\textheight,keepaspectratio]{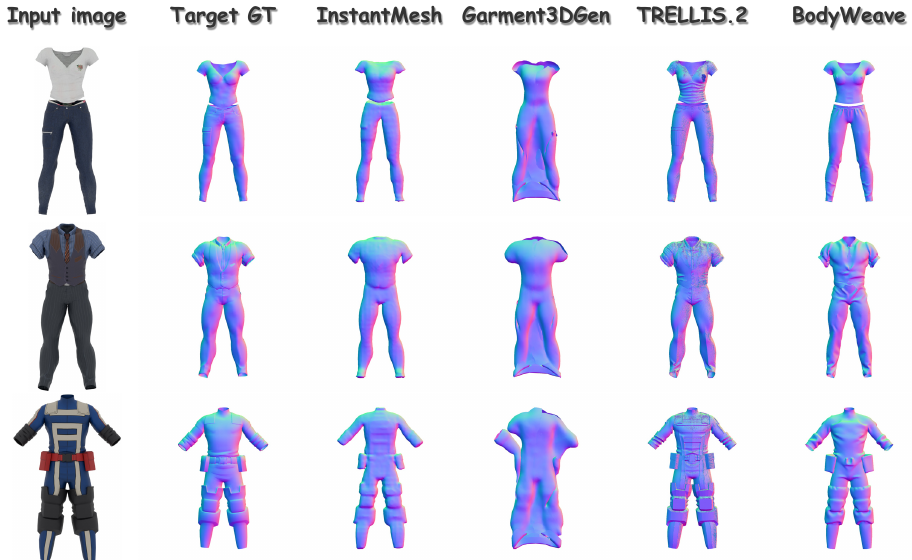}
\caption{Shape reconstruction examples for three artist-authored garments. Normal renderings show silhouette and surface detail.}
\label{fig:supp-controls}
\end{minipage}\hfill
\begin{minipage}[t]{.49\linewidth}
\vspace{0pt}
\captionsetup{type=figure}
\centering
\includegraphics[width=\linewidth,height=.20\textheight,keepaspectratio]{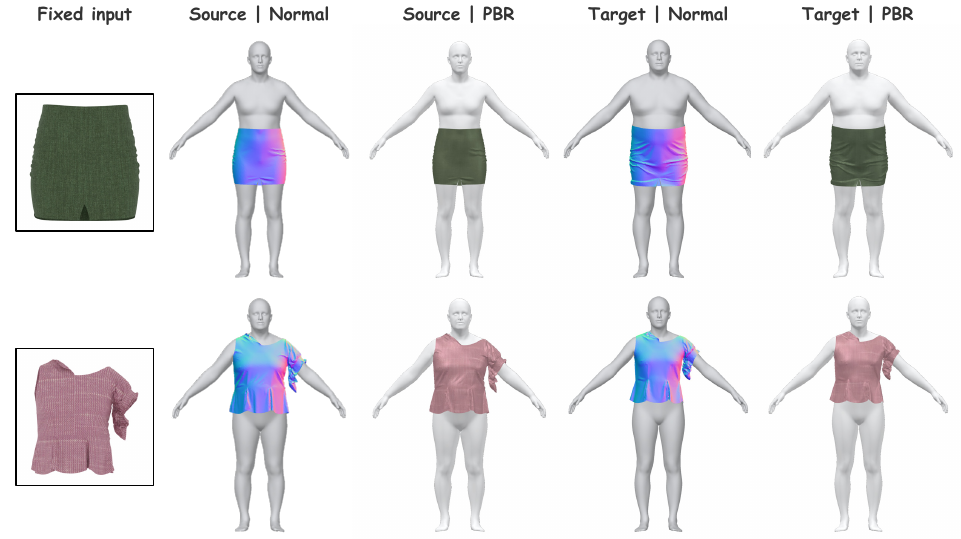}
\caption{Garment adaptation under fixed source images and changing bodies. Normal and PBR renderings show fit and appearance.}
\label{fig:supp-adaptation}
\end{minipage}
\end{center}

\subsection{Generation on Randomly Assigned Bodies}
Randomly assigned bodies test design preservation and body compatibility beyond paired target-garment references.
Silhouette IoU and DINO measure design similarity after independent centering and normalization, while penetration and clearance deviation assess fit in the body frame.
Clearance deviation measures departures from regional garment--body distance ranges calibrated on 64 disjoint validation designs (\cref{sec:supp-evaluation}).
Each method receives the reference information supported by its native input interface (\cref{sec:supp-protocol}).
BodyWeave combines the highest silhouette IoU and DINO similarity with the lowest P95 penetration depth and clearance deviation in \cref{tab:s5}, indicating consistent design preservation and body compatibility across the evaluated outputs.
\begin{center}
\begin{minipage}{\linewidth}
\captionsetup{type=table}
\footnotesize\setlength{\tabcolsep}{3pt}
\caption{Random-body generation. Quality metrics average over evaluable outputs. Penetration and success are percentages, P95 is in mm, and clearance deviation is dimensionless.}\label{tab:s5}
\centering
\begin{tabular*}{\linewidth}{@{}>{\raggedright\arraybackslash}p{\dimexpr0.28000000\linewidth-\tabcolsep\relax}*{6}{>{\raggedleft\arraybackslash}p{\dimexpr0.12000000\linewidth-2\tabcolsep\relax}}@{\hspace{\tabcolsep}}}
\toprule
\textbf{Method} & \textbf{IoU$\uparrow$} & \textbf{DINO$\uparrow$} & \textbf{Pen.$\downarrow$} & \textbf{P95$\downarrow$} & \textbf{Clr. dev.$\downarrow$} & \textbf{Succ.\%$\uparrow$} \\
\midrule
Garment Particles~\cite{nakayama2026garment} & \metriccell{tablepoor!60!tablemid}{0.539} & \metriccell{tablepoor!100!tablemid}{0.291} & \metriccell{tablepoor!20!tablemid}{0.501} & \metriccell{tablepoor!20!tablemid}{16.84} & \metriccell{tablegood!20!tablemid}{0.218} & \metriccell{tablepoor!100!tablemid}{75.00\%} \\
AIpparel~\cite{nakayama2025aipparel} & \metriccell{tablegood!20!tablemid}{0.595} & \metriccell{tablegood!60!tablemid}{\underline{0.341}} & \metriccell{tablegood!60!tablemid}{\underline{0.259}} & \metriccell{tablegood!20!tablemid}{10.40} & \metriccell{tablepoor!20!tablemid}{0.512} & \metriccell{tablepoor!33!tablemid}{82.81\%} \\
SewingLDM~\cite{liu2025multimodal} & \metriccell{tablepoor!20!tablemid}{0.543} & \metriccell{tablegood!20!tablemid}{0.336} & \metriccell{tablegood!100!tablemid}{\textbf{0.106}} & \metriccell{tablegood!60!tablemid}{\underline{9.92}} & \metriccell{tablegood!60!tablemid}{\underline{0.157}} & \metriccell{tablegood!33!tablemid}{92.19\%} \\
Design2Cloth~\cite{zheng2024design2cloth} & \metriccell{tablepoor!100!tablemid}{0.525} & \metriccell{tablepoor!60!tablemid}{0.292} & \metriccell{tablepoor!60!tablemid}{8.364} & \metriccell{tablepoor!60!tablemid}{28.04} & \metriccell{tablepoor!60!tablemid}{0.596} & \metriccell{tablegood!100!tablemid}{\textbf{100.00\%}} \\
Garment3DGen~\cite{sarafianos2025garment3dgen} & \metriccell{tablegood!60!tablemid}{\underline{0.741}} & \metriccell{tablepoor!20!tablemid}{0.295} & \metriccell{tablepoor!100!tablemid}{25.713} & \metriccell{tablepoor!100!tablemid}{66.46} & \metriccell{tablepoor!100!tablemid}{1.305} & \metriccell{tablegood!100!tablemid}{\textbf{100.00\%}} \\
Ours & \metriccell{tablegood!100!tablemid}{\textbf{0.870}} & \metriccell{tablegood!100!tablemid}{\textbf{0.436}} & \metriccell{tablegood!20!tablemid}{0.357} & \metriccell{tablegood!100!tablemid}{\textbf{5.89}} & \metriccell{tablegood!100!tablemid}{\textbf{0.056}} & \metriccell{tablegood!100!tablemid}{\textbf{100.00\%}} \\
\bottomrule
\end{tabular*}
\end{minipage}
\end{center}

\subsection{Appearance Fidelity in Held-Out Views}
\label{sec:supp-appearance}
Shared-geometry evaluation isolates appearance fidelity from shape reconstruction.
All methods receive the front image and the same GT geometry, with left and right views held out. Renderings share cameras, foreground masks, and reference images (\cref{sec:supp-protocol}).
\Cref{tab:s6} separates input-view reconstruction from held-out-view synthesis and also reports their aggregate.
On held-out views, BodyWeave achieves 22.11\,dB PSNR, 0.783 SSIM, and 0.159 LPIPS, improving all three measures over the compared methods.
These results demonstrate appearance fidelity beyond the input viewpoint under shared geometry and rendering conditions.
\begin{center}
\begin{minipage}{\linewidth}
\captionsetup{type=table}
\small\setlength{\tabcolsep}{4pt}
\caption{Appearance synthesis on shared geometry using the front view, with left and right views held out.}\label{tab:s6}
\centering
\begin{tabular*}{\linewidth}{@{}>{\raggedright\arraybackslash}p{\dimexpr0.36000000\linewidth-\tabcolsep\relax}*{4}{>{\raggedleft\arraybackslash}p{\dimexpr0.16000000\linewidth-2\tabcolsep\relax}}@{\hspace{\tabcolsep}}}
\toprule
\textbf{Method} & \textbf{PSNR$\uparrow$} & \textbf{SSIM$\uparrow$} & \textbf{LPIPS$\downarrow$} & \textbf{DINO$\uparrow$} \\
\midrule
\multicolumn{5}{@{}l@{}}{\emph{All views}} \\
FabricDiffusion~\cite{zhang2024fabricdiffusion} & \metriccell{tablepoor!100!tablemid}{13.91} & \metriccell{tablepoor!100!tablemid}{0.510} & \metriccell{tablepoor!100!tablemid}{0.348} & \metriccell{tablepoor!100!tablemid}{0.460} \\
TRELLIS.2~\cite{xiang2026native} & \metriccell{tablegood!0!tablemid}{\underline{15.28}} & \metriccell{tablegood!0!tablemid}{\underline{0.660}} & \metriccell{tablegood!0!tablemid}{\underline{0.212}} & \metriccell{tablegood!100!tablemid}{\textbf{0.753}} \\
Ours & \metriccell{tablegood!100!tablemid}{\textbf{22.42}} & \metriccell{tablegood!100!tablemid}{\textbf{0.790}} & \metriccell{tablegood!100!tablemid}{\textbf{0.143}} & \metriccell{tablegood!0!tablemid}{\underline{0.739}} \\
\multicolumn{5}{@{}l@{}}{\emph{Input view}} \\
FabricDiffusion~\cite{zhang2024fabricdiffusion} & \metriccell{tablepoor!100!tablemid}{13.62} & \metriccell{tablepoor!100!tablemid}{0.495} & \metriccell{tablepoor!100!tablemid}{0.324} & \metriccell{tablepoor!100!tablemid}{0.501} \\
TRELLIS.2~\cite{xiang2026native} & \metriccell{tablegood!0!tablemid}{\underline{15.35}} & \metriccell{tablegood!0!tablemid}{\underline{0.658}} & \metriccell{tablegood!0!tablemid}{\underline{0.189}} & \metriccell{tablegood!0!tablemid}{\underline{0.855}} \\
Ours & \metriccell{tablegood!100!tablemid}{\textbf{23.03}} & \metriccell{tablegood!100!tablemid}{\textbf{0.802}} & \metriccell{tablegood!100!tablemid}{\textbf{0.112}} & \metriccell{tablegood!100!tablemid}{\textbf{0.869}} \\
\multicolumn{5}{@{}l@{}}{\emph{Held-out views}} \\
FabricDiffusion~\cite{zhang2024fabricdiffusion} & \metriccell{tablepoor!100!tablemid}{14.06} & \metriccell{tablepoor!100!tablemid}{0.518} & \metriccell{tablepoor!100!tablemid}{0.360} & \metriccell{tablepoor!100!tablemid}{0.439} \\
TRELLIS.2~\cite{xiang2026native} & \metriccell{tablegood!0!tablemid}{\underline{15.25}} & \metriccell{tablegood!0!tablemid}{\underline{0.660}} & \metriccell{tablegood!0!tablemid}{\underline{0.223}} & \metriccell{tablegood!100!tablemid}{\textbf{0.702}} \\
Ours & \metriccell{tablegood!100!tablemid}{\textbf{22.11}} & \metriccell{tablegood!100!tablemid}{\textbf{0.783}} & \metriccell{tablegood!100!tablemid}{\textbf{0.159}} & \metriccell{tablegood!0!tablemid}{\underline{0.674}} \\
\bottomrule
\end{tabular*}
\end{minipage}
\end{center}

\Cref{fig:supp-material-channels} shows rendered appearance and the predicted base-color, metallic, and roughness channels on shared geometry.
The quantitative comparison measures rendered appearance; the channel visualizations illustrate the generated material representation.
\begin{center}
\begin{minipage}{\linewidth}
\captionsetup{type=figure}
\centering
\includegraphics[width=\linewidth]{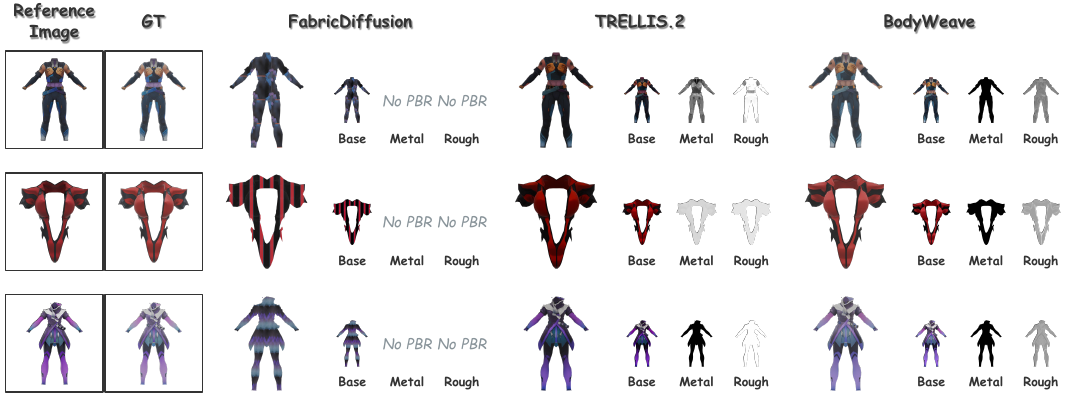}
\caption{Appearance and material maps on shared geometry. ``No PBR'' marks unavailable channels.}
\label{fig:supp-material-channels}
\end{minipage}
\end{center}

\section{Component Ablations and Penetration Severity}
\label{sec:supp-ablation}
\subsection{Global and Local Body Conditioning}
GBA and LBA provide complementary control over global garment placement and local body compatibility.
\Cref{tab:s9} compares the full model with variants without GBA or LBA on 64 GarmentWeave3D reconstruction cases and 16 cross-body tasks.
Without GBA, cross-body CD reaches 275.50\,mm and clearance error reaches 194.10\,mm, indicating substantial errors in body-relative placement.
Without LBA, cross-body CD remains at 9.94\,mm, but penetration rises to 9.338\%, revealing the importance of local body features for surface compatibility.
Combining both modules achieves 8.78\,mm CD and 0.204\% penetration, coupling accurate garment geometry with close adaptation to the target body.
The reconstruction results exhibit the same pattern, supporting the complementary roles of global and local conditioning across both tasks.

\begin{center}
\begin{minipage}{\linewidth}
\captionsetup{type=table}
\centering
\small
\setlength{\tabcolsep}{4pt}
\caption{Effects of global body conditioning, local body conditioning, and surface penetration supervision on reconstruction and cross-body adaptation. Distances are in mm; F@10 and Pen. are percentages. Bold marks the best results within each group at the reported precision. Evaluation uses 50K surface samples per mesh.}
\label{tab:s9}
\begin{tabular*}{\linewidth}{@{\extracolsep{\fill}}lrrrrrrr@{}}
\toprule
Setting & CD$\downarrow$ & F@10$\uparrow$ & NC$\uparrow$ & Pen.$\downarrow$ & P95$\downarrow$ & Clr.$\downarrow$ & Infl.$\downarrow$ \\
\midrule
\multicolumn{8}{@{}l@{}}{\textbf{Matched-body reconstruction on GarmentWeave3D}} \\
\multicolumn{8}{@{}l@{}}{\emph{Body conditioning}} \\
w/o GBA & 279.20 & 1.42 & 0.702 & 0.711 & 28.79 & -- & -- \\
w/o LBA & 7.11 & 81.09 & 0.917 & 4.201 & 5.34 & -- & -- \\
Full model & \textbf{6.36} & \textbf{83.68} & \textbf{0.920} & \textbf{0.080} & \textbf{2.58} & -- & -- \\
\midrule
\multicolumn{8}{@{}l@{}}{\emph{Surface penetration supervision}} \\
w/o penetration loss & 6.43 & 83.29 & 0.916 & 0.092 & \textbf{2.35} & -- & -- \\
w/ penetration loss & \textbf{6.38} & \textbf{83.63} & \textbf{0.917} & \textbf{0.076} & 2.38 & -- & -- \\

\midrule
\multicolumn{8}{@{}l@{}}{\textbf{Cross-body adaptation}} \\
\multicolumn{8}{@{}l@{}}{\emph{Body conditioning}} \\
w/o GBA & 275.50 & 1.72 & 0.700 & 0.311 & -- & 194.10 & 193.40 \\
w/o LBA & 9.94 & 70.43 & 0.879 & 9.338 & -- & 6.11 & 3.00 \\
Full model & \textbf{8.78} & \textbf{75.55} & \textbf{0.884} & \textbf{0.204} & -- & \textbf{4.85} & \textbf{2.64} \\
\midrule
\multicolumn{8}{@{}l@{}}{\emph{Surface penetration supervision}} \\
w/o penetration loss & 8.88 & 75.41 & 0.884 & 0.231 & -- & \textbf{4.90} & 2.66 \\
w/ penetration loss & \textbf{8.80} & \textbf{75.69} & \textbf{0.885} & \textbf{0.190} & -- & \textbf{4.90} & \textbf{2.64} \\

\bottomrule
\end{tabular*}
\end{minipage}
\end{center}

\subsection{Surface Supervision and Intersection Severity}
Surface supervision reduces both the depth and spatial extent of garment--body intersections.
We compare models fine-tuned with and without the penetration loss for the same number of steps, measuring intersection severity with 200K surface samples per mesh (\cref{tab:supp-penetration-severity}).
Mean penetration depth, sampled maximum depth, and the surface area affected by deep intersections decrease in both reconstruction and cross-body adaptation.
For cross-body adaptation, sampled maximum depth falls from 6.60 to 4.54\,mm, a 31.1\% reduction, while CD remains nearly unchanged at 8.88 versus 8.80\,mm (\cref{tab:s9}).
\Cref{fig:supp-penetration} shows five examples with close-ups of garment--body intersections.

\begin{figure}[!htbp]
\centering
\begin{minipage}[t]{.48\linewidth}
\vspace{0pt}
\captionsetup{type=table}
\centering
\small
\setlength{\tabcolsep}{4pt}
\renewcommand{\arraystretch}{1.0}
\parbox[t][52pt]{\linewidth}{\caption{Penetration depth and surface area using 200K samples per mesh. Bold marks the lower mean.}\label{tab:supp-penetration-severity}}
\par\smallskip
\begin{tabular*}{\linewidth}{@{\extracolsep{\fill}}lrr@{}}
\toprule
Metric$\downarrow$ & w/o loss & w/ loss \\
\midrule
\multicolumn{3}{@{}l@{}}{\emph{GarmentWeave3D reconstruction}} \\
Mean depth (mm) & 0.0042 & \textbf{0.0033} \\
RMS depth (mm) & 0.071 & \textbf{0.064} \\
Penetration (\%) & 0.092 & \textbf{0.078} \\
Max. depth (mm) & 3.23 & \textbf{3.00} \\
Area $>1$\,mm (\%) & 0.059 & \textbf{0.047} \\
Area $>2$\,mm (\%) & 0.045 & \textbf{0.034} \\
Area $>5$\,mm (\%) & 0.027 & \textbf{0.019} \\
Area $>10$\,mm (\%) & 0.012 & \textbf{0.010} \\
\midrule
\multicolumn{3}{@{}l@{}}{\emph{Cross-body adaptation}} \\
Mean depth (mm) & 0.0156 & \textbf{0.0111} \\
RMS depth (mm) & 0.204 & \textbf{0.151} \\
Penetration (\%) & 0.229 & \textbf{0.186} \\
Max. depth (mm) & 6.60 & \textbf{4.54} \\
Area $>1$\,mm (\%) & 0.195 & \textbf{0.150} \\
Area $>2$\,mm (\%) & 0.168 & \textbf{0.128} \\
Area $>5$\,mm (\%) & 0.113 & \textbf{0.086} \\
Area $>10$\,mm (\%) & 0.063 & \textbf{0.042} \\
\bottomrule
\end{tabular*}
\end{minipage}\hfill
\begin{minipage}[t]{.48\linewidth}
\vspace{0pt}
\captionsetup{type=figure}
\centering
\parbox[t][52pt]{\linewidth}{\caption{Penetration supervision examples. Insets show inside-body depth on a shared 0--10\,mm scale.}\label{fig:supp-penetration}}
\par\smallskip
\includegraphics[width=\linewidth,height=.38\textheight,keepaspectratio]{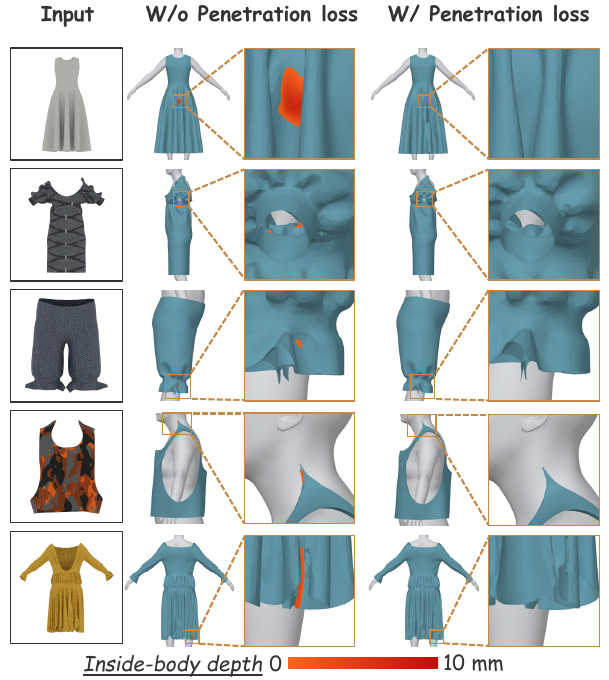}
\end{minipage}
\end{figure}

\section{Evaluation Metrics and Baseline Protocols}
\label{sec:supp-protocols}
\subsection{Evaluation Metrics and Aggregation}
\label{sec:supp-evaluation}
\paragraph{Geometric fidelity.}
We evaluate surface reconstruction with non-squared symmetric Chamfer distance (CD), F-score at a 10\,mm threshold (F@10), and absolute normal consistency (NC).
Geometry metrics use 50K area-uniform surface samples per mesh.
For normalized-shape evaluation, independent AABB centering and maximum-side normalization remove global translation and scale, yielding dimensionless CD and an F-score threshold of 1\% of the normalized unit.

\paragraph{Body penetration.}
Let $d_B(p)$ denote the target-body signed distance in millimeters, positive outside the body, and define penetration depth as $h(p)=\max(-d_B(p),0)$.
Penetration rate is the percentage of sampled points with $d_B(p)<0$. P95 is the 95th percentile of depth among penetrating points, set to zero for meshes without penetration.
Penetration severity is evaluated using 200K area-uniform surface samples per mesh.
Mean and RMS depth are computed over all sampled points, including zeros outside the body; Max. is the sampled maximum depth, and Area $>t$ is the percentage of points with $h(p)>t$.
Each statistic is computed per mesh before averaging across garments.

\paragraph{Garment--body spacing.}
Clearance error measures agreement with the reference garment's distance to the body, while inflation isolates excess outward spacing.
Let $P$ and $Q$ be area-uniform surface samples from the predicted and GT garments, respectively.
Define nearest-neighbor correspondences $\nu_Q(p)=\arg\min_{q\in Q}\|p-q\|_2$ and $\nu_P(q)=\arg\min_{p\in P}\|q-p\|_2$, and signed spacing differences
\begin{equation}
\Delta_P(p)=d_B(p)-d_B(\nu_Q(p)),\qquad
\Delta_Q(q)=d_B(\nu_P(q))-d_B(q).
\end{equation}
The two metrics are
\begin{align}
\mathrm{Clr.}&=\frac{1}{2}\left(\frac{1}{|P|}\sum_{p\in P}|\Delta_P(p)|+\frac{1}{|Q|}\sum_{q\in Q}|\Delta_Q(q)|\right),\\
\mathrm{Infl.}&=\frac{1}{2}\left(\frac{1}{|P|}\sum_{p\in P}[\Delta_P(p)]_++\frac{1}{|Q|}\sum_{q\in Q}[\Delta_Q(q)]_+\right),
\end{align}
where $[a]_+=\max(a,0)$. Both metrics use symmetric surface correspondences and are reported in millimeters.

\paragraph{Body response and direction.}
Body-response metrics assess whether generated garments change with the wearer in the same way as the reference garments.
Let $m(G)\in\mathbb R^6$ collect garment measurements in millimeters, with $G_s,G_t$ denoting source and target GT garments and $\hat G_s,\hat G_t$ their predictions.
For the frozen-placement control, $m$ comprises three axis extents and three coordinates of the area-weighted surface centroid.
With $\Delta m=m(G_t)-m(G_s)$ and $\Delta\hat m=m(\hat G_t)-m(\hat G_s)$, response error is
\begin{equation}
E_{\mathrm{resp}}=\frac{1}{6}\sum_{j=1}^{6}|\Delta\hat m_j-\Delta m_j|.
\end{equation}
Response direction (R-dir.) compares predicted source-to-target changes using a 2\,mm dead zone:
\begin{equation}
s_\epsilon(a)=\begin{cases}0,&|a|\leq\epsilon,\\ \operatorname{sign}(a),&|a|>\epsilon,\end{cases}
\qquad
\mathrm{R\text{-}dir.}=\frac{100}{6}\sum_{j=1}^{6}\mathbf 1[s_\epsilon(\Delta\hat m_j)=s_\epsilon(\Delta m_j)],\quad \epsilon=2\,\mathrm{mm}.
\end{equation}
Target-change direction (T-dir.) in \cref{tab:s2,tab:s2b} uses the same sign-agreement formula with $\Delta\hat m=m(\hat G_t)-m(G_s)$, taking the source GT as the common reference.
For T-dir., $m$ comprises the three axis extents, area-weighted centroid height, and top and bottom heights. Both direction metrics compare predicted changes against the corresponding GT change $m(G_t)-m(G_s)$.

\paragraph{Clearance deviation without paired GT.}
For randomly assigned bodies, clearance deviation measures compatibility with garment--body spacing observed in a disjoint validation set.
We partition each garment into 24 regions using six equal-height bands of its AABB and four azimuth sectors about its horizontal AABB center.
For each region $r$, let $\ell_r$ and $u_r$ be the 5th and 95th percentiles of body signed distances pooled from 64 validation designs disjoint from evaluation.
The same regional bounds are used for all methods. For a predicted surface point $p$ in region $r(p)$, clearance deviation is
\begin{equation}
D_{\mathrm{clr}}=\frac{1}{|P|}\sum_{p\in P}
\frac{[\ell_{r(p)}-d_B(p)]_+ + [d_B(p)-u_{r(p)}]_+}
{\max(u_{r(p)}-\ell_{r(p)},\varepsilon)},\qquad \varepsilon=10^{-6}\,\mathrm{mm}.
\end{equation}
This dimensionless score is zero within the regional distance range and increases with deviations outside it. Averaging over points weights regions by sampled surface area; empty prediction regions contribute no samples.

\paragraph{Image and appearance fidelity.}
For random-body generation, silhouette IoU and DINO similarity assess consistency with the source images after independent centering and normalization of the predictions. Body penetration and clearance are evaluated in the original body frame.
For appearance synthesis, all methods share GT geometry and rendering conditions. PSNR uses mean squared RGB error within the GT foreground mask, and SSIM averages the local similarity map over that mask.
LPIPS uses AlexNet, while DINO similarity is the cosine similarity of L2-normalized final-layer CLS features from DINOv2-Large.
Both perceptual metrics use the GT foreground bounding box with eight pixels of padding and a common gray background outside the mask.
Input-view scores use the front view, held-out scores average the left and right views, and all-view scores average all three views with equal weight.

\subsection{Baseline Interfaces and Evaluation Protocol}
\label{sec:supp-protocol}
Each baseline retains its native input representation and uses the body-adaptation procedure appropriate to the evaluated task.

\paragraph{Simulation-based methods.}
Garment Particles~\cite{nakayama2026garment}, AIpparel~\cite{nakayama2025aipparel}, and SewingLDM~\cite{liu2025multimodal} use their native generation interfaces followed by simulation on the target body, without target-garment alignment.
AIpparel does not predict the attachment labels required by the shared simulator, so its attachment constraints differ from those of the other simulation-based methods.
For SewingLDM, Qwen2.5-VL-3B-Instruct~\cite{bai2025qwen25vltechnicalreport} generates an image-specific caption from each source front RGB image using a shared instruction template.
The template is fixed across samples; the resulting descriptions depend on the visible garment.
We use deterministic decoding with a 72-token output limit.
SewingLDM's original T5 encoder embeds the captions alongside a front Canny sketch and its supported body-PCA condition.
For Artist-Created bodies outside the PCA representation, generation uses the PCA origin before simulation on the actual target body.
\begin{center}
\begingroup
\setlength{\fboxsep}{10pt}
\setlength{\fboxrule}{0.5pt}
\fcolorbox{black!25}{black!3}{%
\begin{minipage}{\dimexpr\linewidth-2\fboxsep-2\fboxrule\relax}
\small\raggedright
{\bfseries SewingLDM caption-generation prompt}\par
\vspace{4pt}
{\color{black!25}\hrule height 0.4pt}
\vspace{7pt}
\setlength{\parskip}{6pt}
\setlength{\parindent}{0pt}
You are generating a garment description for the SewingLDM sewing-pattern model.

Inspect only the visible FRONT of the isolated garment image. Ignore the black background, the mannequin or body if present, lighting, shadows, and rendering artifacts. Do not describe the image or the background. Do not infer the back side or details that are not visible.

Return exactly ONE line in this format and nothing else:\par
{\ttfamily\footnotesize Upper garment: <concise visible sewing/fashion description>; Lower garment: <concise visible sewing/fashion description>}\par

Use SewingLDM-style terms such as garment category, silhouette, sleeve or strap type, neckline, waist, visible length, color, material, and pattern. Keep EACH field under 18 words. Do not add sub-labels such as ``Neckline:'', ``Material:'', or ``Pattern:'', and do not add an explanation or a list. Use ``unknown'' when a field cannot be determined. If the garment is one-piece, describe it in the upper garment field and write ``same one-piece garment'' in the lower garment field. Never mention the image background, mannequin, person, or image itself.
\end{minipage}}
\endgroup
\end{center}

\paragraph{Direct 3D methods.}
Design2Cloth~\cite{zheng2024design2cloth} uses a garment mask and an adapted SMPL body. Target-pose transfer uses linear blend skinning, with weights interpolated from the four nearest rest-body vertices by inverse squared distance, without garment-GT fitting.
Garment3DGen~\cite{sarafianos2025garment3dgen} uses a front image and a fixed longsleeve template. Matched-body reconstruction retains its InstantMesh~\cite{xu2024instantmesh} initialization and applies GT-assisted rotation, isotropic scaling, and translation.
For cross-body adaptation, a top, dress, or skirt category annotated from the source image determines placement: garment height is $\alpha H$, vertical center is $y_{\min}+\beta H$, and horizontal AABB centers are aligned, where $H$ and $y_{\min}$ are the target body's height and lower vertical bound.
We use $(\alpha,\beta)=(0.38,0.72)$ for tops, $(0.72,0.55)$ for dresses, and $(0.40,0.36)$ for skirts.
TRELLIS.2~\cite{xiang2026native} retains its image-only input, with shape normalization defined in \cref{sec:supp-evaluation}.
The fixed-geometry control reuses one 1024-cascade output from the source front image for both bodies, applying scale 0.5 and translating its vertex centroid to each body's AABB center in the native z-up frame.
The separate BodyWeave intervention holds the model and sampling seed fixed while changing the body condition. The mismatched body is selected from the other seven designs to approximate the correct target's source-relative distance in bust, waist, hips, and height measurements.

Appearance comparisons with TRELLIS.2 and FabricDiffusion~\cite{zhang2024fabricdiffusion} use shared GT geometry, cameras, and foreground masks, rendered at $512^2$ with maximum-side scale 0.85.
All methods receive only the front image. Left and right views are held out for evaluation.

\section{Representation and Training Details}
\label{sec:method-details}
\subsection{Representation and Image Conditions}
The geometry cascade uses a shared body-global coordinate convention at all resolutions.
BodyWeave uses garment-only orthographic front, left, and right views rendered with a shared three-view scale and a maximum garment AABB extent of 0.85 in the rendering frame.
SS/LR consume these images at $512^2$ resolution and HR/PBR at $1024^2$, without additional bounding-box or random resized crops.
All 3D supervision remains in the body frame.
The sparse-structure VAE encodes garment occupancy and the clean target body into eight-channel $N_{\mathrm{SS}}^3$ latents, with $N_{\mathrm{SS}}=16$.
The SS denoiser has 30 transformer blocks with 1536-dimensional hidden features and independent residual body adapters.
Each GBA uses a 256-dimensional hidden space, and SS training updates the adapters while freezing the pretrained denoiser.
The local body encoder projects 32-channel sparse features to 512-dimensional attention features, with 32 nearest body tokens per garment query.
LR and HR use $32^3$ and $64^3$ latent grids and jointly train the SLat backbone, body projection, and LBA modules.
Both GBA and LBA have zero-initialized output projections.
HR is initialized from LR weights with reset optimizer.
The appearance model is fine-tuned from the 1024-resolution TRELLIS.2 texture-flow model~\cite{xiang2026native}.

\subsection{Training and Surface Supervision}
\paragraph{Optimization.}
\Cref{tab:supp-training} summarizes the stage-specific settings.
All stages use AdamW with $\beta_1=0.9$, $\beta_2=0.95$, weight decay 0.01, and $\epsilon=10^{-8}$.
Learning rates remain constant, without warm-up or decay, and model weights use an exponential moving average with decay 0.9999.
Training uses BF16 mixed precision. SDF queries are evaluated in FP32.
\begin{center}
\begin{minipage}{\linewidth}
\captionsetup{type=table}
\centering\small
\caption{Training configuration by stage. All stages use flow matching; LR and HR additionally use surface penetration supervision.}
\label{tab:supp-training}
\begin{tabular*}{\linewidth}{@{\extracolsep{\fill}}lcccc@{}}
\toprule
Setting & SS & LR & HR & PBR \\
\midrule
Learning rate & $10^{-4}$ & $10^{-4}$ & $2\times10^{-5}$ & $2\times10^{-5}$ \\
Accumulation splits & 8 & 2 & 1 & 1 \\
Optimization steps & 20K & 10K & 65K & 58K \\
\bottomrule
\end{tabular*}
\end{minipage}
\end{center}

\paragraph{Flow matching.}
The flow path uses $\sigma=10^{-5}$, with logit-normal time sampling for SS (mean 1 and standard deviation 1 before the sigmoid) and uniform sampling for LR and HR.
Image and body conditions are independently dropped with probability 0.1 for classifier-free guidance.
Body dropout bypasses GBA and LBA, and the Surface Penetration Loss excludes points from examples whose body condition is dropped.
The frozen DINOv3 image encoder processes all three views, whose feature sequences are concatenated for image cross-attention.
The PBR model concatenates 32-channel appearance and 32-channel shape latents at each occupied coordinate.

\paragraph{Surface supervision.}
Surface supervision uses the predicted clean latent from the same flow-matching forward pass, followed by latent denormalization and differentiable vertex decoding.
We decode every example in the micro-batch and pool the resulting garment vertices. If the pool exceeds 20,000 vertices, we sample 20,000 uniformly without replacement; otherwise, all vertices are used.
Each sampled vertex queries its own target body's SDF, and vertices from body-dropped examples receive zero loss weight.
We precompute body SDFs by voxelizing and filling the body on a shared global lattice at nominal resolutions 512 and 1024 for LR and HR, respectively.
Interior occupancy is obtained by intersecting the filled intervals along the three coordinate axes and retaining the voxelized surface. Euclidean distance transforms inside and outside the occupancy yield signed distances with a half-voxel boundary correction, positive outside and negative inside.
Each field is stored as a body-local crop of the global lattice with a 0.03-model-unit margin. Differentiable trilinear interpolation evaluates distances at decoded vertices; points outside the crop use the nonnegative boundary value plus distance to the crop AABB.
The body SDF uses the shared model coordinate unit, with one unit corresponding to 2.32 meters.
We set the safety margin to $m=0.001$ model units, equivalent to $2.32\,\mathrm{mm}$, and use $\eta=0.05$ and $\delta=10^{-8}$ for the Surface Penetration Loss.

\subsection{Inference}
We generate garments with the SS--LR--HR cascade, followed by geometry-conditioned PBR synthesis.
SS predicts occupied garment locations on the $32^3$ LR latent grid. The shape decoder upsamples the generated LR representation to determine occupied locations on the $64^3$ HR latent grid, where HR shape generation starts from independent Gaussian noise.
The final shape and appearance latents are decoded at a spatial resolution of 1024.

The default inference configuration uses 12 Euler integration steps for each of SS, LR, HR, and PBR, with $\sigma_{\min}=10^{-5}$ and random seed 42.
Starting from a uniform grid $u_i=1-i/12$, we rescale flow time as $\tau_i=a u_i/[1+(a-1)u_i]$, with $a=5$ for SS and $a=3$ for LR, HR, and PBR.
SS and both shape stages use image classifier-free guidance of strength 7.5 for $\tau\in[0.6,1.0]$; outside this interval, sampling uses the conditional prediction.
Guidance rescaling interpolates between the guided clean-latent estimate and its standard-deviation-matched counterpart, with coefficients 0.7 for SS and 0.5 for LR/HR.
The target-body condition is retained in both image-conditioned and image-unconditioned predictions throughout geometry sampling.
PBR uses guidance strength 1, conditioning on the generated HR shape and reference images.

\clearpage
\section{Extended Qualitative Results}
\label{sec:supp-qualitative}
\Cref{fig:supp-reconstruction,fig:supp-reconstruction-artist} present additional reconstruction examples on GarmentWeave3D and Artist-Created Garments, showing garment silhouettes, folds, and fine structures.
\Cref{fig:supp-appearance,fig:supp-appearance-more} compare rendered appearance on shared geometry and visualize the available material channels.
Baseline processing follows \cref{sec:supp-protocol}.

\begin{figure}[!htbp]
\centering
\includegraphics[width=\linewidth,height=.76\textheight,keepaspectratio]{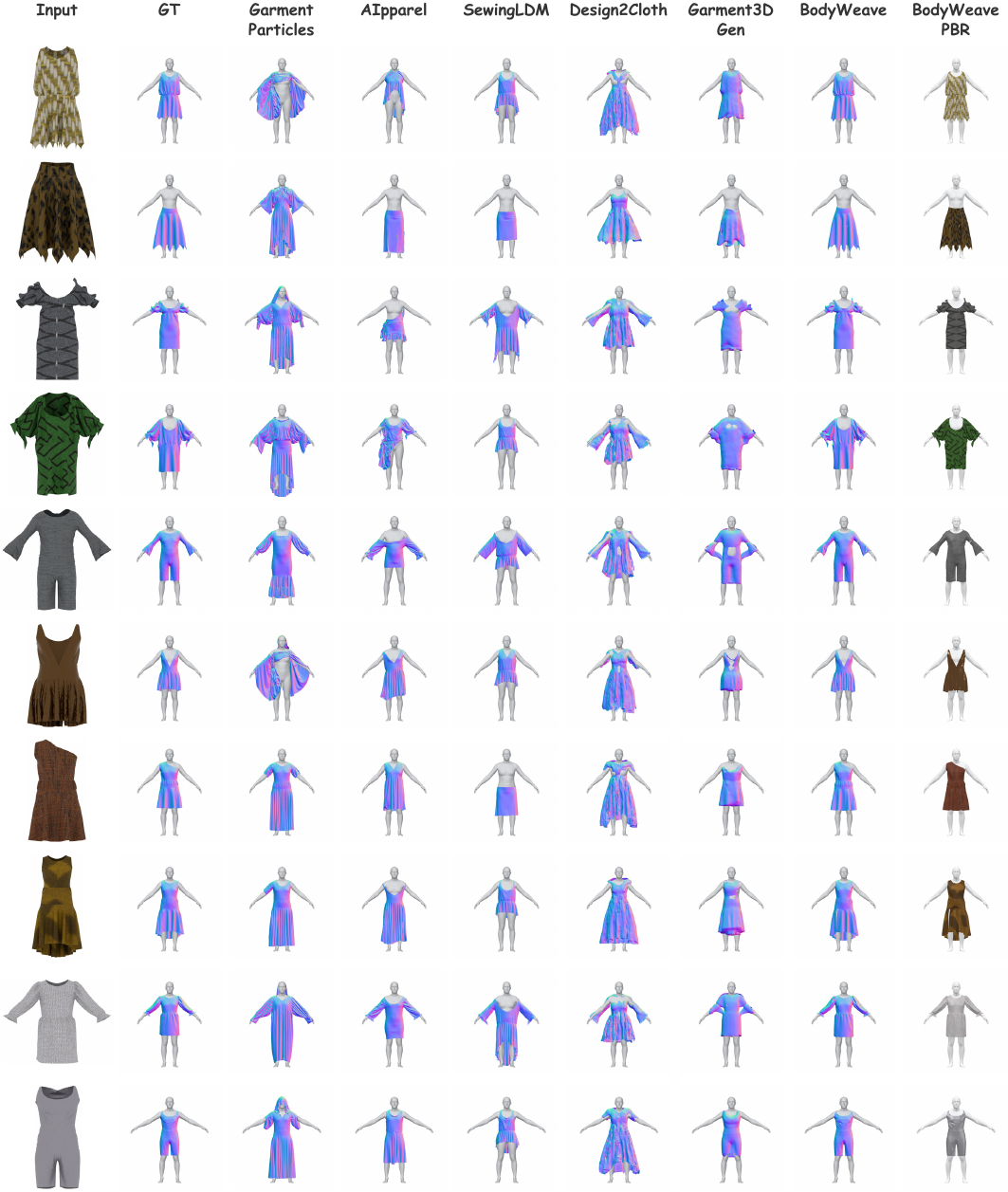}
\caption{Additional geometric reconstruction examples on GarmentWeave3D. Normal renderings show silhouettes and surface detail.}
\label{fig:supp-reconstruction}
\end{figure}
\begin{figure}[p]
\centering
\includegraphics[width=\linewidth,height=.83\textheight,keepaspectratio]{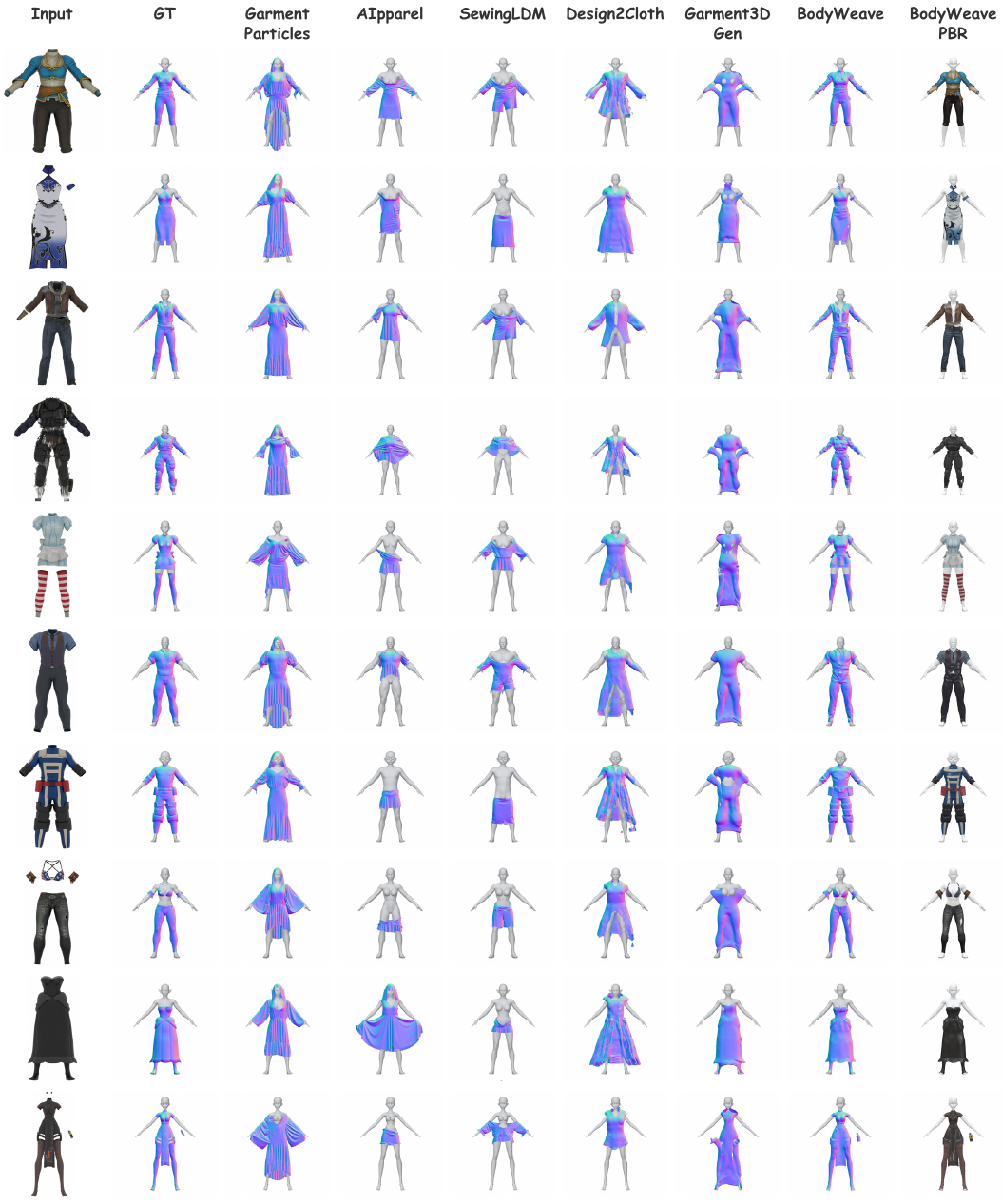}
\caption{Geometric reconstruction examples on artist-authored garments with varied silhouettes and structures.}
\label{fig:supp-reconstruction-artist}
\end{figure}
\begin{figure}[p]
\centering
\includegraphics[width=\linewidth,height=.83\textheight,keepaspectratio]{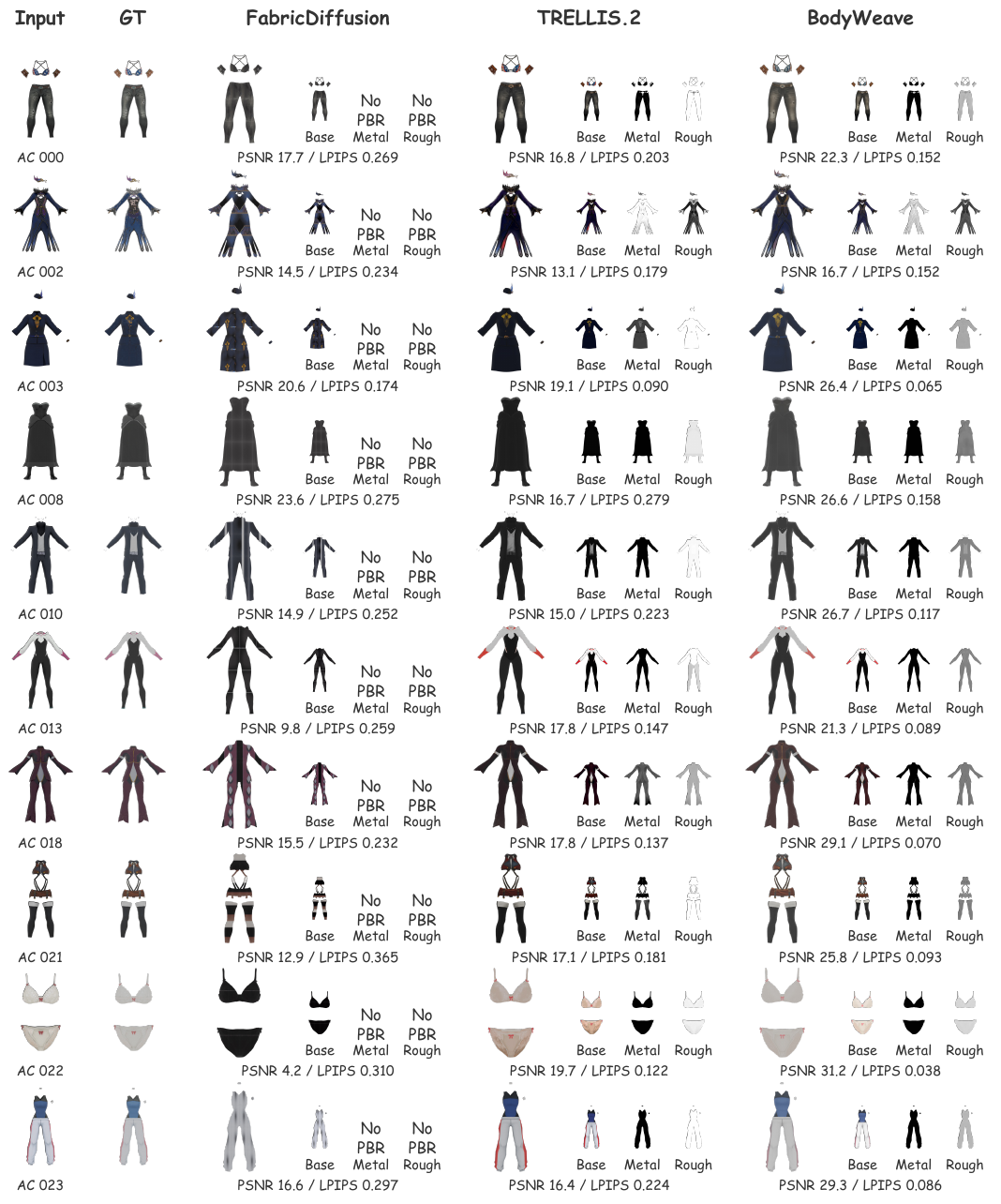}
\caption{Appearance and material maps on shared geometry. Insets show available base-color, metallic, and roughness channels.}
\label{fig:supp-appearance}
\end{figure}
\begin{figure}[p]
\centering
\includegraphics[width=\linewidth,height=.83\textheight,keepaspectratio]{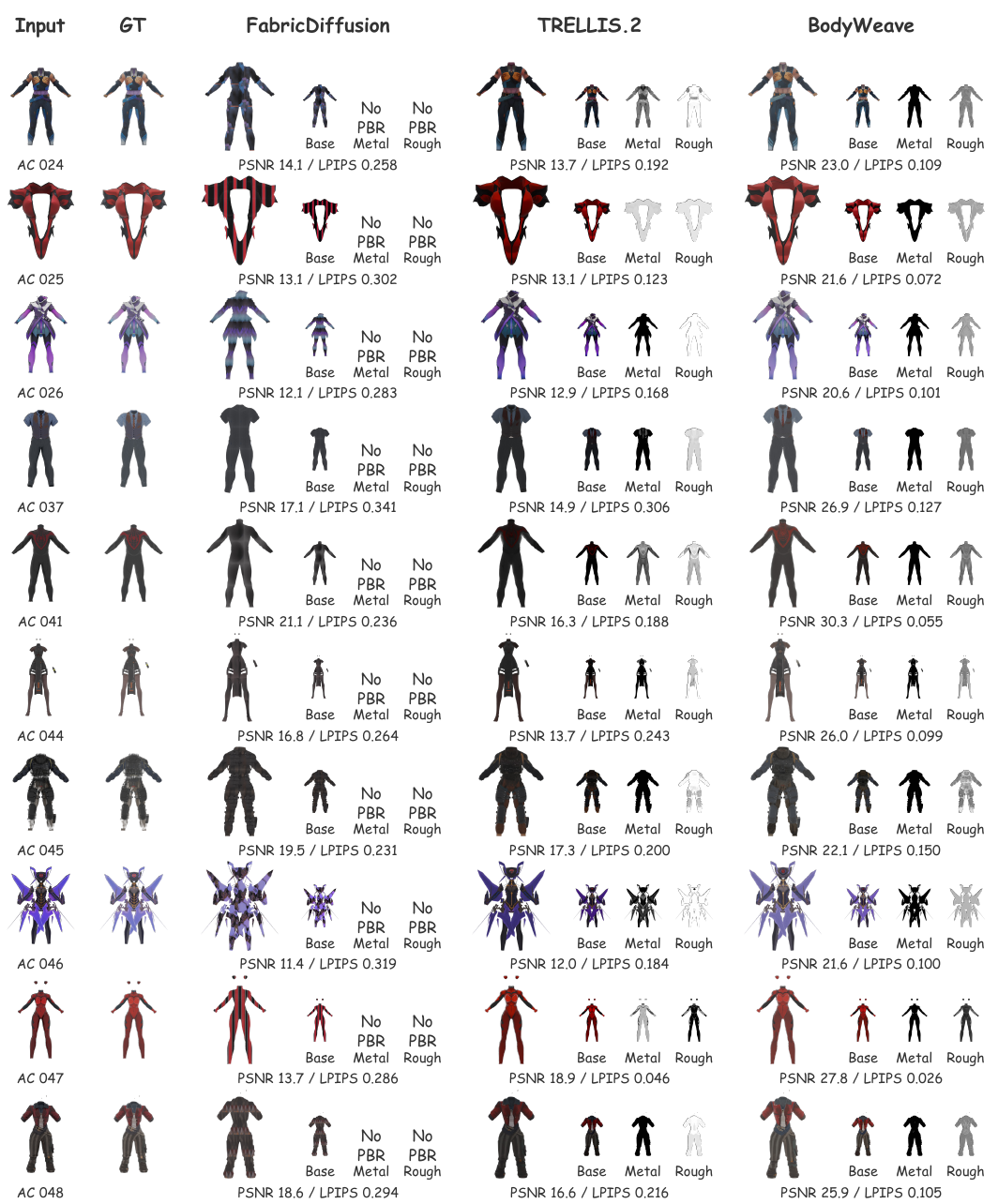}
\caption{Appearance and material maps on artist-authored garments.}
\label{fig:supp-appearance-more}
\end{figure}
\clearpage
\endgroup

\end{document}